\documentclass[pdflatex,sn-vancouver-ay]{sn-jnl}

\usepackage{graphicx}%
\usepackage{multirow}%
\usepackage{amsmath,amssymb,amsfonts}%
\usepackage{amsthm}%
\usepackage{mathrsfs}%
\usepackage[title]{appendix}%
\usepackage{xcolor}%
\usepackage{textcomp}%
\usepackage{manyfoot}%
\usepackage{booktabs}%
\usepackage{longtable}%
\usepackage{tabularx}%
\usepackage[ruled,vlined,linesnumbered]{algorithm2e}
\usepackage{algpseudocode}%
\usepackage{listings}%

\usepackage{graphicx}%
\usepackage{multirow}%
\usepackage{subcaption}
\usepackage{enumitem}
\usepackage{threeparttable}

\usepackage[table]{xcolor}
\usepackage{array}
\usepackage{placeins}

 \usepackage{tikz}
 \usetikzlibrary{
 arrows.meta,
 backgrounds,
 fit,
 positioning
}

\definecolor{Ink}{HTML}{172033}
\definecolor{Muted}{HTML}{667085}
\definecolor{Line}{HTML}{CBD5E1}

\definecolor{Agg}{HTML}{2563EB}
\definecolor{AggSoft}{HTML}{EFF6FF}

\definecolor{NonAgg}{HTML}{7C3AED}
\definecolor{NonAggSoft}{HTML}{F5F3FF}

\definecolor{Target}{HTML}{059669}
\definecolor{TargetSoft}{HTML}{ECFDF5}

\newcommand{\killpunct}[1]{}

\newif\ifdraft

\newif\ifcomment
\commenttrue

 \newtheorem{revcomment}{Reviewer Comment \#}[section]

\newcommand{\grayed}[1]{{\cellcolor[gray]{.7}}}

\theoremstyle{thmstyleone}%
\theoremstyle{thmstyletwo}%
\newtheorem{example}{Example}%

\theoremstyle{thmstylethree}%

\newcommand{\component}[1]{\operatorname{\mathsf{#1}}}

\newcommand{\result}[2]{#1\,{\tiny (#2)}}

\begin{document}

\title[A Framework for Graph Node Symmetric Quantification under
Dataset Shift]{A Framework for Graph Node Symmetric Quantification
under Dataset Shift}


\author[1]{\fnm{Alessio}
\sur{Micheli}}\email{alessio.micheli@unipi.it}

\author*[2]{\fnm{Alejandro} \sur{Moreo}}\email{alejandro.moreo@cnr.it}

\author[1]{\fnm{Marco} \sur{Podda}}\email{marco.podda@unipi.it}

\author[2]{\fnm{Fabrizio}
\sur{Sebastiani}}\email{fabrizio.sebastiani@cnr.it}

\author[1]{\fnm{Domenico}
\sur{Tortorella}}\email{domenico.tortorella@unipi.it}

\affil[1]{\orgdiv{Dipartimento di Informatica}, \orgname{Università di
Pisa}, \orgaddress{\street{Largo Bruno Pontecorvo 3}, \city{Pisa},
\postcode{56127}, \country{Italy}}}

\affil[2]{\orgdiv{Istituto di Scienza e Tecnologie dell'Informazione},
\orgname{Consiglio Nazionale delle Ricerche}, \orgaddress{\street{Via
Giuseppe Moruzzi 1}, \city{Pisa}, \postcode{56124}, \country{Italy}}}


\abstract{\textit{Graph node quantification} is the supervised
learning task of estimating the prevalence of different classes within
an arbitrary sample of unlabelled nodes from a graph, subject to some
form of dataset shift. Existing approaches to graph node
quantification typically follow the standard aggregative paradigm,
i.e., they use labelled nodes from the graph to train an intermediate
node classifier, and obtain an estimated class distribution for the
sample by aggregating the classifier's predictions for the individual
nodes and then correcting the result via some adjustment strategy that
accounts for dataset shift. In this work we propose a framework for
graph node quantification that does away with both node-level
supervision, since its training phase only requires node samples
labelled by their class distribution, and node-level predictions,
since it estimates class prevalence values directly, without the need
of any intermediate node classifier. The framework thus allows
\textit{symmetric} quantification, since (unlike in standard
``asymmetric'' quantification methods) the labelled items used for
training and the unlabelled items on which predictions must be issued
are of the same type (i.e., node samples). This framework (that we dub
GSQ, for \underline{G}raph Node \underline{S}ymmetric
\underline{Q}uantification) allows building pipelines consisting of a
node encoder, an embedding pooler, a prevalence predictor, and an
optional post-hoc scaling step. These components can be instantiated
by choosing from different trainable and reservoir-based encoders,
different pooling operators, and different prediction objectives. We
evaluate the GSQ framework on binary and multi-class datasets with
different levels of homophily, under prior probability shift and under
structural covariate shift.
The experiments show that GSQ remains competitive with strong
aggregative quantifiers under prior probability shift and achieve
lower test error in seven of the eighth comparisons under structural
covariate shift, with substantially tied performance in the remaining
case. Component-wise comparisons further show that the preferred
architecture depends on graph homophily and the sampling
protocol. These results support GSQ as a practical alternative,
especially in cases when supervision is available only at the sample
level and/or when individual node predictions are undesirable.
We make available all the code and datasets that are necessary to
replicate our results.}

\keywords{Quantification, Supervised prevalence estimation, Class
distribution estimation, Graph node quantification, Graph learning,
Dataset shift, Non-aggregative quantification methods, Symmetric
quantification methods}



\maketitle



\section{Introduction}
\label{sec:introduction}

\noindent
\textit{Quantification}~\citep{Esuli:2023os, Gonzalez:2017it,
Moreo:2026gd} is the supervised learning task of estimating the class
prevalence values of a sample (i.e., set) of unlabelled datapoints
affected by some form of dataset shift.\footnote{For reasons that we
will explain in \S\ref{sec:node-quantification}, the samples that
quantification addresses are best viewed as \textit{multisets} (rather
than plain sets) of datapoints. For the moment being, we may take this
distinction as inessential.} This problem was first explicitly
formulated (although without an explicit mention of dataset shift)
about 20 years ago~\citep{Forman:2005fk}, and for the first 10 years
it encountered limited interest on the part of the scientific
community. However, this interest has progressively increased since
then, also due to an increased awareness of the impact of dataset
shift \citep{Moreno-Torres:2012ay, Quinonero:2009kl} on the
performance of machine-learned models.

While most quantification literature addresses ``flat'' data, a few
works~\citep{Damke:2025cr, Damke:2025qf, Micheli:2025ez, Milli:2015mz,
Tang:2010uq} have tackled the case (dubbed \textit{graph node
quantification} -- aka ``graph quantification'' or ``network
quantification'') in which the (training and unlabelled) datapoints
are nodes in a graph. As is well-known from classification
(quantification's ``sibling'' task), the fact that datapoints are
interconnected is an important source of information, and classifiers
capable of exploiting this information routinely outperform
classifiers that treat the graph as a flat set of
datapoints~\citep{Sen:2008fk}; the same is true of quantification, as
has been shown in the above-cited papers.

When learning (either a classifier or a quantifier) in a
networked-data context, we usually assume that the training data and
the unlabelled data on which inference must be performed are nodes of
the \textit{same} graph, and that the structure of the entire graph
(i.e., which nodes are connected to which other nodes) \textit{and}
the covariates of all the nodes in the graph are known at training
time; in other words, one difference between learning from flat data
and learning from networked data is that this latter is usually
tackled as a \textit{transductive} learning
task~\citep{Gammerman:1998ay}.

In this paper we contribute to the literature on graph node
quantification by presenting a new transductive framework for
estimating class prevalence values in samples of unlabelled graph
nodes. A \textit{symmetric quantification method} is one that, unlike
most quantification methods, (i) does not require labelled nodes for
training, and can be directly trained from node samples labelled with
their class distribution, and (ii) does not issue predictions for
individual nodes, but directly predicts the class distribution of the
unlabelled sample. Symmetric quantification methods are thus called
since (unlike for the more common ``asymmetric'' ones) the labelled
items used for training and the unlabelled items on which predictions
must be issued are of the same type (here: node samples).
Symmetric quantification methods are of independent interest, because
(a) they are the full realisation of the well-known \textit{Vapnik's
principle} (\citep{Vapnik98} -- see also~\cite[\S 4.4]{Esuli:2023os}),
according to which we should solve our task (here: quantification)
directly and not by first solving a more general task (here:
classification), and (b) in a number of contexts (especially ones
where privacy is a concern -- see Example~\ref{ex:example1} in
\S\ref{sec:symmetricvsasymmetric}), training information might be
available at the aggregate (i.e., sample) level but not at the
individual datapoint level, and drawing inferences at the aggregate
level might be desirable but drawing inferences at the individual
level might not~\citep{Fabris:2023pj}.


While our framework (dubbed GSQ, for \underline{G}raph Node
\underline{S}ymmetric \underline{Q}uantification) is not specific to
any type of dataset shift, in our experiments we test it in scenarios
characterised by two different types of such shift, i.e.,
\textit{prior probability shift} (PPS -- an important type of dataset
shift) and \textit{structural covariate shift} (a type of dataset
shift specific to graph settings, and recently illustrated by
\cite{Damke:2025cr, Damke:2025qf}), by adopting experimental protocols
that allow us to inject these types of shift in the data.\footnote{We
call GSQ a ``framework'', rather than a ``method'', since it allows
building pipelines consisting of several cascaded components (a node
encoder, an embedding pooler, a prevalence predictor, and an optional
post-hoc scaling step) where different choices for each such component
are provided.}
Our experiments show that the proposed approach remains competitive
with aggregative methods under prior probability shift and generally
outperforms them when queries are generated through structural
sampling, while preserving the distinctive advantage of being
applicable when individual node labels are unavailable for training or
individual predictions are undesirable.

The present paper extends a previous work of ours
\citep{Micheli:2026xp} in a number of ways, i.e., by (a) tackling also
multi-class (and not only binary) node quantification, (b) doubling
the number of datasets considered, (c) extensively studying and
assessing the core model by varying node encoders, poolers, and
prevalence predictors, while also considering post-hoc prediction
scaling, and finally (d) evaluating quantification accuracy also on
structural covariate shift, alongside the more frequently addressed
prior probability shift.

The rest of the paper is organised as follows. \S\ref{sec:background}
briefly describes the core concepts behind our contribution, while
also introducing mathematical
notation. \S\ref{sec:symmetric-quantification} presents our main
contribution, the GSQ symmetric framework for graph node
quantification, including its components, and how it can be trained
and evaluated. \S\ref{sec:experiments} details the experiments carried
on to assess GSQ's performance under two important subtypes of dataset
shift. \S\ref{sec:results} discusses the results we have obtained, and
assesses (by analysing the results of extensive ablation experiments)
the contribution of the different components of
GSQ. \S\ref{sec:related-work} surveys related work, and
\S\ref{sec:conclusion} concludes, pointing at avenues for future
research. Further details about our experiments are provided in
Appendix~\ref{app:additional-details-experiments}, while
Appendix~\ref{app:additional-results} presents the complete set of
results for the main experiments.

We make available all the code and datasets that are necessary to
replicate our
results.\footnote{\label{foot:code}\url{https://github.com/marcopodda/gsq}}


\section{Problem setting}
\label{sec:background}
\noindent This section introduces background notions on ``flat'',
non-relational quantification (\S\ref{sec:quantification}) and on
graph node quantification (\S\ref{sec:node-quantification}). Following
this, it presents a discussion of symmetric vs.\ asymmetric
quantification (\S\ref{sec:symmetricvsasymmetric}), which motivate our
proposal of a graph node symmetric quantification framework. Finally,
we introduce dataset shift and its main subtypes
(\S\ref{sec:datasetshift}), and protocols for simulating them in
experiments (\S\ref{sec:datasetshiftsimulation}).


\subsection{Quantification}
\label{sec:quantification}
\noindent Let $\mathcal{X}$ be an input space and $\mathcal{Y}$ a
finite set of class labels. Given a non-empty multiset of datapoints
$\sigma=\{(x_i,y_i)\}_{i=1}^{|\sigma|}$ (hereafter: a
\textit{sample})\footnote{In the quantification literature, $\sigma$
is often defined not as a plain set but as a \textit{multiset} (or
\textit{bag}) of datapoints, i.e., $\sigma$ is allowed to contain
multiple copies of the same datapoint; accordingly, the cardinality of
$\sigma$ and sums over $\sigma$ treat each instance as a different
datapoint. Among other things, using multisets allows the experimental
protocols typically used in quantifier evaluation (see
\S\ref{sec:experimental-setup}) to sample with replacement, which can
indeed generate samples containing multiple instances of the same
datapoint. Viewing $\sigma$ as a multiset, instead of a plain set, is
not problematic for quantification methods, since none of them depends
on the elements of the sample being distinct.} with unobserved labels
$y_i\in\mathcal{Y}$,
quantification aims to estimate its class prevalence vector
$\mathbf{p}_\sigma$, whose entries are
\begin{equation}\label{eq:class-prevalence}
  p_\sigma(y)=\frac{1}{|\sigma|}\sum_{i=1}^{|\sigma|}\mathbb{I}[y_i=y]
  \qquad y\in\mathcal{Y}
\end{equation}
%
A quantifier $q$ maps a sample $\sigma$ of unlabelled datapoints
(which we will sometimes refer to as the \textit{query}) to an
estimate $\mathbf{\widehat{p}}_\sigma=q(\sigma)$. Notice that
$\mathbf{\widehat{p}}_\sigma$ must belong to the probability simplex
$\Delta^{|\mathcal{Y}|-1}$, meaning that its entries must satisfy
$\widehat p_\sigma(y)\geq 0$ for every $y\in\mathcal{Y}$ and
$\sum_{y\in\mathcal{Y}}\widehat p_\sigma(y)=1$.

In order to evaluate the accuracy of a quantifier $q$, the estimated
prevalence vector $\mathbf{\widehat{p}}_\sigma=q(\sigma)$ for a sample
$\sigma$ of test datapoints is compared with the true prevalence
vector $\mathbf{p}_\sigma$ using a quantification error measure. The
most satisfactory such measures (see~\citet{Sebastiani:2020qf} for a
discussion) are \textit{Absolute Error} (AE) and \textit{Relative
Absolute Error} (RAE), defined as
\begin{align}
  \operatorname{AE}(\mathbf p_\sigma,\mathbf{\widehat{p}}_\sigma)
  &=\frac{1}{|\mathcal Y|}\sum_{y\in\mathcal Y}
    |p_\sigma(y)-\widehat p_\sigma(y)|\label{eq:mae} \\
  \operatorname{RAE}(\mathbf p_\sigma,\mathbf{\widehat{p}}_\sigma)
  &=\frac{1}{|\mathcal Y|}\sum_{y\in\mathcal Y}
    \frac{|p_\sigma(y)-\widehat p_\sigma(y)|}{p_\sigma(y)+\epsilon}\label{eq:rae} 
\end{align}
where $\epsilon=\frac{1}{2|\sigma|}$ is a smoothing constant. A
measure of the quality of the quantification method is then obtained
by averaging across all test samples.


\subsection{Node quantification}\label{sec:node-quantification}

Let $G=(\mathcal{V},\mathcal{E})$ be an undirected graph with nodes
$\mathcal{V} = \{v_i\}_{i=1}^n$ and edges $\mathcal{E} \subseteq
\mathcal{V} \times \mathcal{V}$ encoded as a symmetric binary
adjacency matrix $\mathbf{A} \in \{0,1\}^{n\times n}$, with entries
$a_{ij} = \mathbb{I}[(v_i,v_j) \in \mathcal{E}]$. A node $v$ has
features (aka ``covariates'') $\mathbf{x}_v \in \mathbb{R}^d$ (with $d
\in \mathbb{N}$). By $\mathbf{X} \in \mathbb{R}^{n \times d}$ we
denote the matrix of node features.

As customary in transductive graph learning, $\mathcal{V}$ is
partitioned into labelled nodes $\mathcal{V}_L$ and unlabelled nodes
$\mathcal{V}_U=\mathcal{V}\setminus\mathcal{V}_L$; the learner
observes $\mathbf{X}$, $\mathbf{A}$, and the labelled pairs $\{(v,
y_v) \mid v \in\mathcal{V}_L, y_v \in \mathcal{Y}\}$, and makes
inferences about nodes in $\mathcal{V}_U$. For the node classification
task, the goal is to predict individual labels $y_v$ for
$v\in\mathcal{V}_U$, while for the node quantification task, which is
the object of this study, the goal is to predict the class prevalence
vector $\mathbf{p}_\sigma$ of a non-empty sample $\sigma$ drawn from
$\mathcal{V}_U$.

The predicted vector $\widehat{\mathbf{p}}_\sigma$ must belong to the
probability simplex $\Delta^{|\mathcal{Y}|-1}$, so its entries are
non-negative and sum to one. We denote a predicted prevalence vector's
entry corresponding to class $y$ by $\widehat p_{\sigma}(y)$.


\subsection{Symmetric node
quantification}\label{sec:symmetricvsasymmetric}
In the literature, quantification has been predominantly tackled as an
\textit{asymmetric} task, and \textit{node} quantification (see
\S\ref{sec:related-work} for an overview) has been exclusively tackled
so. Here, ``asymmetric'' refers to the fact that, in most
quantification methods, supervision comes in the form of labelled
\textit{datapoints}, while quantification requires that predictions
are issued not on individual datapoints but on \textit{samples} of
datapoints; this is in sharp contrast with other supervised learning
tasks (e.g., classification, regression) in which both training data
and unlabelled data are of the same type. In turn, asymmetric
quantification methods can be of the \textit{aggregative} or of the
\textit{non-aggregative} type; both train their models on individual
labelled datapoints, but the former first issue predictions for
individual datapoints (either in the form of class labels or in the
form of posterior probabilities) and then derive an estimated class
distribution from them, while the latter directly estimate a class
distribution without the help of a classifier.

%
\begin{figure}
  \centering \resizebox{\linewidth}{!}{

\begin{tikzpicture}[
  font=\sffamily\small,
  framework box/.style={
    rectangle,
    rounded corners=4pt,
    line width=0.8pt,
    minimum width=.26\linewidth,
    text width=.225\linewidth,
    minimum height=76pt,
    inner xsep=6pt,
    inner ysep=6pt,
    align=center,
    anchor=center,
    text={rgb,255:red,51;green,51;blue,51},
    fill opacity=1,
    text opacity=1
  },
  flow arrow/.style={
    draw={rgb,255:red,102;green,102;blue,102},
    line width=1pt,
    -{Latex[length=4.5pt,width=3.5pt]},
    shorten <=3pt,
    shorten >=3pt
  },
  query arrow/.style={
    draw={rgb,255:red,200;green,140;blue,80},
    line width=1pt,
    -{Latex[length=4.5pt,width=3.5pt]},
    shorten <=3pt,
    shorten >=3pt
  },
  group label/.style={
    inner sep=0pt,
    font=\sffamily\bfseries\small
  }
]

\node[framework box,
  draw={rgb,255:red,105;green,95;blue,150},
  fill={rgb,255:red,239;green,237;blue,247}
] (graph) {
  \textbf{Input graph}\\[9pt]
  $(\mathcal V_L\cup\mathcal V_U,\mathcal E)$\\[6pt]
  $(\mathbf X,\mathbf A)$
};

\node[framework box,right=.06\linewidth of graph,
  draw={rgb,255:red,80;green,140;blue,200},
  fill={rgb,255:red,231;green,239;blue,247}
] (encoder) {
  \textbf{Node encoder}\\[10pt]
  $\mathbf H
    =\component{enc}_{\omega}
      (\mathbf X,\mathbf A)$
};

\node[framework box,right=.06\linewidth of encoder,
  draw={rgb,255:red,80;green,140;blue,200},
  fill={rgb,255:red,231;green,239;blue,247}
] (classifier) {
  \textbf{Node classifier}\\[10pt]
  $\widehat{\mathbf Y}
    =\component{clf}_{\psi}(\mathbf H)$
};

\node[framework box,below=44pt of classifier,
  draw={rgb,255:red,180;green,120;blue,140},
  fill={rgb,255:red,245;green,236;blue,239}
] (quantifier) {
  \textbf{Aggregative quantifier}\\[8pt]
  $\begin{gathered}
    \widehat{\mathbf p}_{\sigma}
      =q(\widehat{\mathbf Y}_{\sigma})\\[3pt]
    \widehat{\mathbf Y}_{\sigma}
      =\{\widehat{\mathbf y}_v\mid v\in\sigma\}
  \end{gathered}$
};

\node[framework box,left=.06\linewidth of quantifier,
  draw={rgb,255:red,200;green,140;blue,80},
  fill={rgb,255:red,247;green,239;blue,231}
] (sample) {
  \textbf{Query sample}\\[10pt]
  $\sigma\subseteq\mathcal V_U$
};

\node[framework box,right=.06\linewidth of quantifier,
  draw={rgb,255:red,120;green,180;blue,140},
  fill={rgb,255:red,236;green,245;blue,239}
] (output) {
  \textbf{Class-prevalence}\\
  \textbf{estimate}\\[10pt]
  $\widehat{\mathbf p}_{\sigma}
    \in\Delta^{|\mathcal Y|-1}$
};

\begin{scope}[on background layer]
  \node[
    rectangle,
    rounded corners=7pt,
    draw=none,
    fill={rgb,255:red,233;green,240;blue,247},
    fit=(encoder)(classifier),
    inner xsep=7pt,
    inner ysep=16pt,
    outer sep=0pt
  ] (graphphase) {};

  \node[
    rectangle,
    rounded corners=7pt,
    draw=none,
    fill={rgb,255:red,247;green,233;blue,237},
    fit=(quantifier),
    inner xsep=7pt,
    inner ysep=16pt,
    outer sep=0pt
  ] (samplephase) {};
\end{scope}

\node[group label,
  anchor=south,
  text={rgb,255:red,55;green,105;blue,155}
] at ([yshift=5pt]graphphase.north) {
  Graph-level inference
};

\node[group label,
  anchor=north,
  text={rgb,255:red,145;green,75;blue,100}
] (samplelabel) at ([yshift=-5pt]samplephase.south) {
  Sample-level inference
};

\draw[flow arrow]
  (graph.east) -- (encoder.west);

\draw[flow arrow]
  (encoder.east) -- (classifier.west);

\draw[flow arrow]
  (classifier.south) -- (quantifier.north);

\draw[query arrow]
  (sample.east) -- (quantifier.west);

\draw[flow arrow]
  (quantifier.east) -- (output.west);

\path ([yshift=-18pt]samplelabel.south)
  node[inner sep=0pt,outer sep=0pt] {};

\end{tikzpicture}}
 \caption{Inference with a trained 
 aggregative node quantification pipeline. The encoder maps the input graph to node embeddings, using which a classifier is first trained from labelled nodes and then applied to issue predictions for the unlabelled nodes. For a query sample $\sigma \subseteq \mathcal{V}_U$, the aggregative quantifier aggregates the corresponding predictions and corrects for dataset shift to produce a class-prevalence estimate. The shaded regions distinguish graph-level from sample-level inference. 
 }
 \label{fig:aggregative-framework}
\end{figure}
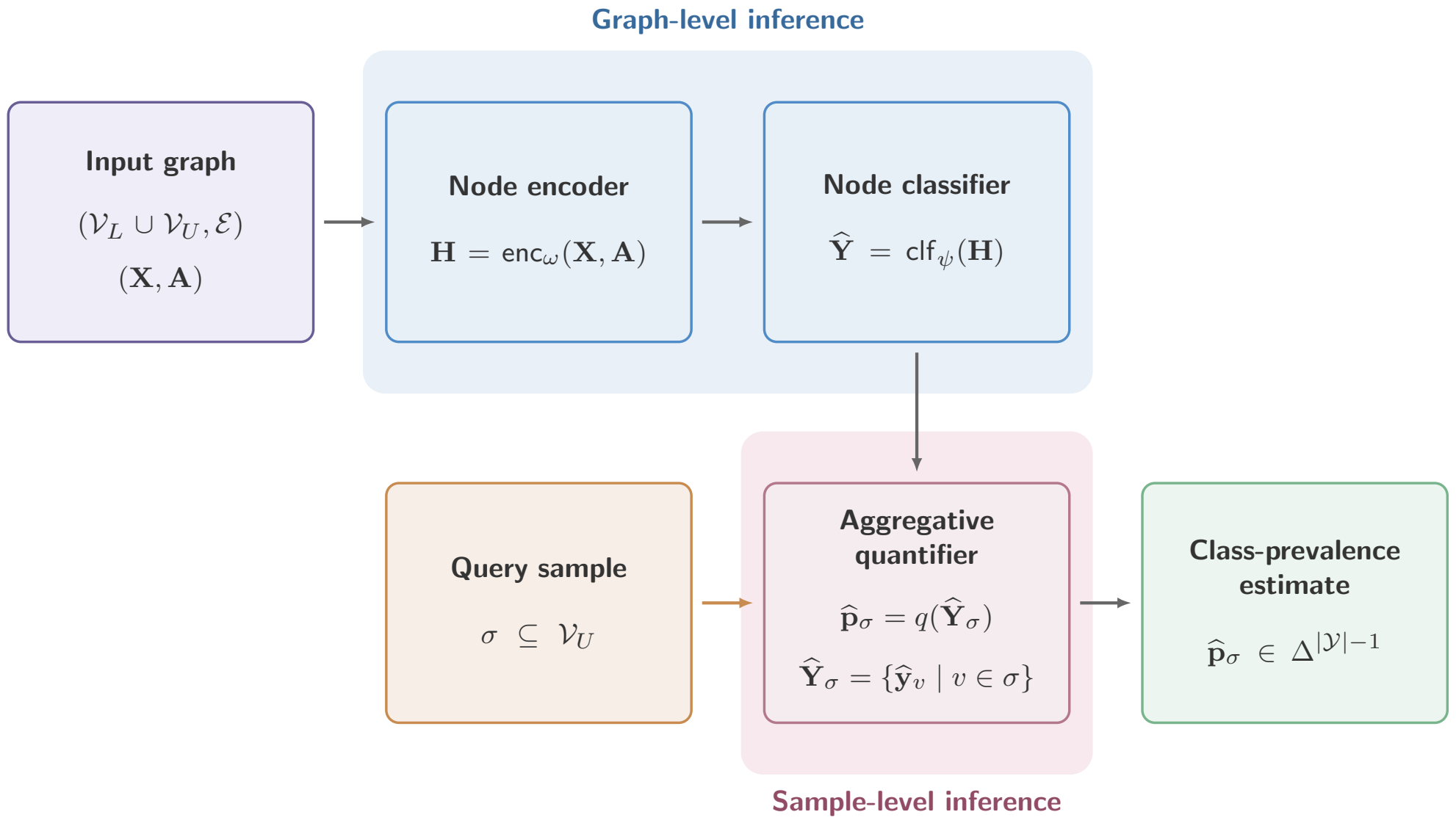

Accordingly, the aggregative (and asymmetric) node quantification
pipeline, exemplified in Figure~\ref{fig:aggregative-framework},
starts with a parametrised node encoder $\component{enc}_{\omega}$,
typically based on message passing \citep{gilmer_neural_2017}. The
encoder ingests a graph's node feature matrix $\mathbf{X}$ and
adjacency matrix $\mathbf{A}$ to compute node embeddings $\mathbf{H}
\in \mathbb{R}^{n \times d_h}$, for some $d_h \in \mathbb{N}$ (usually
chosen via model selection). The node embeddings are passed to a
classifier, typically a linear softmax head or an MLP, which produces
a matrix of class probabilities
$\widehat{\mathbf{Y}}=\component{clf}_{\psi}(\mathbf{H})$. Each row
$\widehat{\mathbf{y}}_v\in\Delta^{|\mathcal{Y}|-1}$ contains the
predicted class probabilities for node $v$. The quantifier $q$ then
aggregates the predictions associated with the nodes in the sample,
usually applying a correction for dataset shift to obtain the
prevalence estimate
$\widehat{\mathbf{p}}_\sigma=q(\widehat{\mathbf{Y}}_\sigma)$, where
$\widehat{\mathbf{Y}}_\sigma$ contains one prediction row per
occurrence in $\sigma$. Methods requiring hard labels instead use
$\widehat y_v=\arg\max_{y\in\mathcal{Y}}\,\widehat{\mathbf{y}}_v(y)$.


Asymmetric (aggregative) node quantifiers have shown good performance
on binary and multi-class tasks across different levels of graph
homophily \citep{Micheli:2025ez}. However,
%
asymmetric methods also require the availability of individual
labelled nodes for training and of predictions for individual nodes.
When the above cannot be obtained, as in applications where
privacy-by-design constraints hold, the use of these methods may thus
be infeasible; the following is an example.

\begin{example}
  \label{ex:example1}
  Suppose our nodes $v_i$ are voters of an administrative election,
  each represented via covariates such as age, income, etc. Suppose
  our edges $(v_i,v_j)$ represent kinship, node labels $y_i$ are the
  party the voters have chosen, and samples $\sigma$ are the sets of
  all voters that belong to a given municipality. Suppose the
  elections are held in different moments in time for different
  regions of the country (with a region consisting of a set of
  municipalities), so that the regions for which the elections are
  held first provide the training data (``training regions'') and the
  other provide the unlabelled data (``test regions'').

  A predictor can thus be trained using data from the training regions
  and then be deployed to predict results for the test regions. We are
  only interested in predictions at the aggregate level (i.e.,
  estimates of the prevalence of the different parties in a given
  region), so the prediction is one of \textit{quantification} rather
  than one of classification. Additionally, it is neither desirable
  nor necessary to issue predictions for the individual datapoints
  (i.e., voters), so the quantifier should be a
  \textit{non-aggregative} one. Finally, since individual votes are
  secret, it is implausible that the training data may come in the
  form of labelled datapoints, while it is perfectly plausible that it
  can come in the form of labelled samples (i.e., aggregate results
  for entire municipalities from the training regions); the
  quantification method should thus be not just non-aggregative, but
  \textit{symmetric}. \qed
\end{example}

\noindent These limitations of graph node asymmetric quantification
methods motivate investigating their symmetric counterparts, which we
do in this paper.


\subsubsection{Symmetric scenarios and symmetric methods}
\label{sec:scenariosandmethods}
\noindent Before moving on, however, let us remark a difference
between symmetric quantification (SQ) \textit{scenarios} (i.e.,
real-world applications in which labels for training are available
only at the aggregate level and in which predictions at the individual
level are undesirable -- see Example~\ref{ex:example1}) and SQ
\textit{methods}. An SQ scenario requires an SQ method, but an SQ
method can operate in any quantification scenarios, even ones where
individual labelled datapoints are available for training; other
things being equal, SQ methods are thus advantageous with respect to
asymmetric ones.

However, for an SQ method, SQ scenarios are, other things being equal,
more difficult to tackle than asymmetric ones. The reason is that,
when individual labelled datapoints are available for training, there
is a combinatorial number of labelled samples that can be generated
for them, while the labelled samples that are available in truly SQ
scenarios are usually limited in number. This is not just an issue of
quantity, though, since the availability of individual labelled
datapoints for training allows choosing which samples to generate for
training, and one can generate the samples (via an appropriate sample
generation protocol -- see \S\ref{sec:experimental-protocols}) that
best suits the type of shift one expects to affect the data.


\subsection{Dataset shift}
\label{sec:datasetshift}
\noindent Quantification methods commonly assume the presence of some
form of \textit{dataset shift}, whereby the joint distribution of
inputs $X$ and labels $Y$ varies between training and test data, i.e.,
$$P_{\mathrm{tr}}(X,Y) \neq P_{\mathrm{te}}(X,Y)$$
with $X$ and $Y$ random variables ranging on $\mathcal{X}$ and
$\mathcal{Y}$, respectively. In order to analyse dataset shift it is
useful to consider two alternative factorisations of $P(X,Y)$, i.e.,
\begin{align}
  \label{eq:causalfactorisation}
  P(X,Y) = & \ P(Y|X)P(X) \\
  \label{eq:anti-causalfactorisation}
  P(X,Y) = & \ P(X|Y)P(Y)
\end{align}
%

\noindent and to introduce the distinction between causal learning
problems and anti-causal learning
problems~\citep{Scholkopf:2012je}. \textit{Causal learning} problems,
aka ``$X\rightarrow Y$ problems''~\citep{Fawcett:2005fk}, are tasks in
which we infer a phenomenon $Y$ from its \textit{causes}; an example
is avalanche prediction, in which we predict whether an avalanche will
fall ($Y$) from physical and meteorological variables ($X$) that
influence the formation of avalanches, such as steepness of the
terrain, recent snowfalls, air temperature, etc. \textit{Anti-causal
learning} problems, aka ``$Y\rightarrow X$
problems''~\citep{Fawcett:2005fk}, are instead tasks in which we infer
a phenomenon $Y$ not from its causes but from its \textit{symptoms},
or \textit{effects}; an example is disease prediction from symptoms,
in which we infer the presence of a given disease ($Y$) from symptoms
($X$) such as nausea, dizziness, fatigue, and the like.

In causal learning problems it is natural to use Equation
\ref{eq:causalfactorisation}, since $P(Y|X)$ makes explicit the causal
dependence of $Y$ on $X$. In these problems, a common type of shift is
\textit{covariate shift} (CS), defined as the case in which $P(Y|X)$
is invariant across training and test data but $P(X)$ is not. An
example of covariate shift is when we train an avalanche predictor on
data from Swiss mountains and we want to deploy it on data from Nepali
mountains, which are characterised by a radically different
distribution of physical and meteorological variables, but where the
causal dependence of avalanches from these variables is unchanged
(since it is dictated by the laws of physics). In causal learning
problems, when $P(Y|X)$ is instead \textit{not} invariant, we are
instead in the presence of \textit{concept shift}, a type of shift
usually considered intractable (unless some amount of labelled data
from $P_{\mathrm{te}}(X,Y)$ is available), since the relationship
between $X$ and $Y$ that allows learning breaks down.

Conversely, in anti-causal learning problems it is natural to use
Equation \ref{eq:anti-causalfactorisation}, since $P(X|Y)$ makes
explicit the causal dependence of $X$ on $Y$. In these problems, a
common type of shift is \textit{prior probability shift} (PPS -- aka
\textit{label shift}), defined as the case in which $P(X|Y)$ is
invariant across training and test data but $P(Y)$ is not. An example
of label shift is when we train our disease prediction system on data
from a non-pandemic scenario and we want to deploy it on data from a
pandemic one, since these two scenarios are characterised by vastly
different label distributions, but where the causal dependence of the
symptoms from the disease is unchanged (since it is dictated by
physiology). In anti-causal learning problems, when $P(X|Y)$ is
instead \textit{not} invariant, we are again in the presence of a
(different) form of concept shift, which is considered as intractable
as the other form we have discussed above.

Most quantification research papers deal with PPS (see
\citep{Esuli:2023os} for a survey), while only a few individual such
efforts~\citep{Card:2018pb, Gonzalez:2024cs, Tasche:2022hh,
Tasche:2023um} deal with covariate shift.
While CS and PPS are by far the most frequently addressed types of
shift in ``generic'' data settings, \citet{Damke:2025cr, Damke:2025qf}
recently discuss a type of shift specific to graph settings called
\textit{structural covariate shift} (SCS), a type of shift informed by
the topology (or ``structure'') of the graph. A typical example arises
when labelled data are available from a certain region of the graph
and predictions must be issued for a different region; for instance,
in a graph where nodes represent cities and edges represent railway
connections, data may be available, say, for Swedish cities but not
for Norwegian ones. SCS is a type of covariate shift, since it is
determined by the link structure of the graph, and links belong to the
input of the problem (the covariates) and not to its output (the
labels).



\subsection{Simulating dataset
shift}\label{sec:datasetshiftsimulation}
A quantifier trained on labelled data drawn from
$P_{\mathrm{tr}}(X,Y)$ receives an unlabelled sample $\sigma$ drawn
from $P_{\mathrm{te}}(X,Y)$ and estimates its class prevalence vector
$\mathbf{p}_\sigma$. The assumptions incorporated in the different
quantification methods about the type of shift at play determine which
properties learned from the training distribution can be exploited
when estimating class prevalence values for the test sample. For
example, methods designed for PPS exploit the train/test invariance of
the conditional $P(X|Y)$.
The assumptions made about the type of shift at play therefore
determine the conditions under which a quantifier can transfer
information from the labelled data to the unlabelled query population.

In the experimental evaluation of a quantifier, if we want to measure
the robustness of a quantifier to a certain type of shift, it is
important that the test data used in the experimentation displays
exactly this type of shift. Since it is not always easy to obtain
enough datasets that display the desired type of shift,
the experimental evaluation of quantifiers usually resorts to
simulating it via a specific simulation \textit{protocol}, which
specifies how to sample the available dataset in order to generate
samples of datapoints that exhibit the desired type of shift. Quite
naturally, the same protocol that is used for sampling the test set
can also be used for sampling the validation set, so as to provide
samples of datapoints that exhibit the type of shift we want the
system to be robust to, and that can be used in the model selection
phase. We will discuss the protocols relevant to our experimentation
in \S\ref{sec:experimental-protocols}.

\section{Symmetric graph node quantification}
\label{sec:symmetric-quantification}
In this section we present our main contribution: GSQ, a symmetric
(hence, non-aggregative) framework for node quantification. We
formulate this problem as a simplex-valued regression task in which a
model learns to predict class prevalence values from a representation
of the test sample, without assigning individual class labels as an
intermediate step.
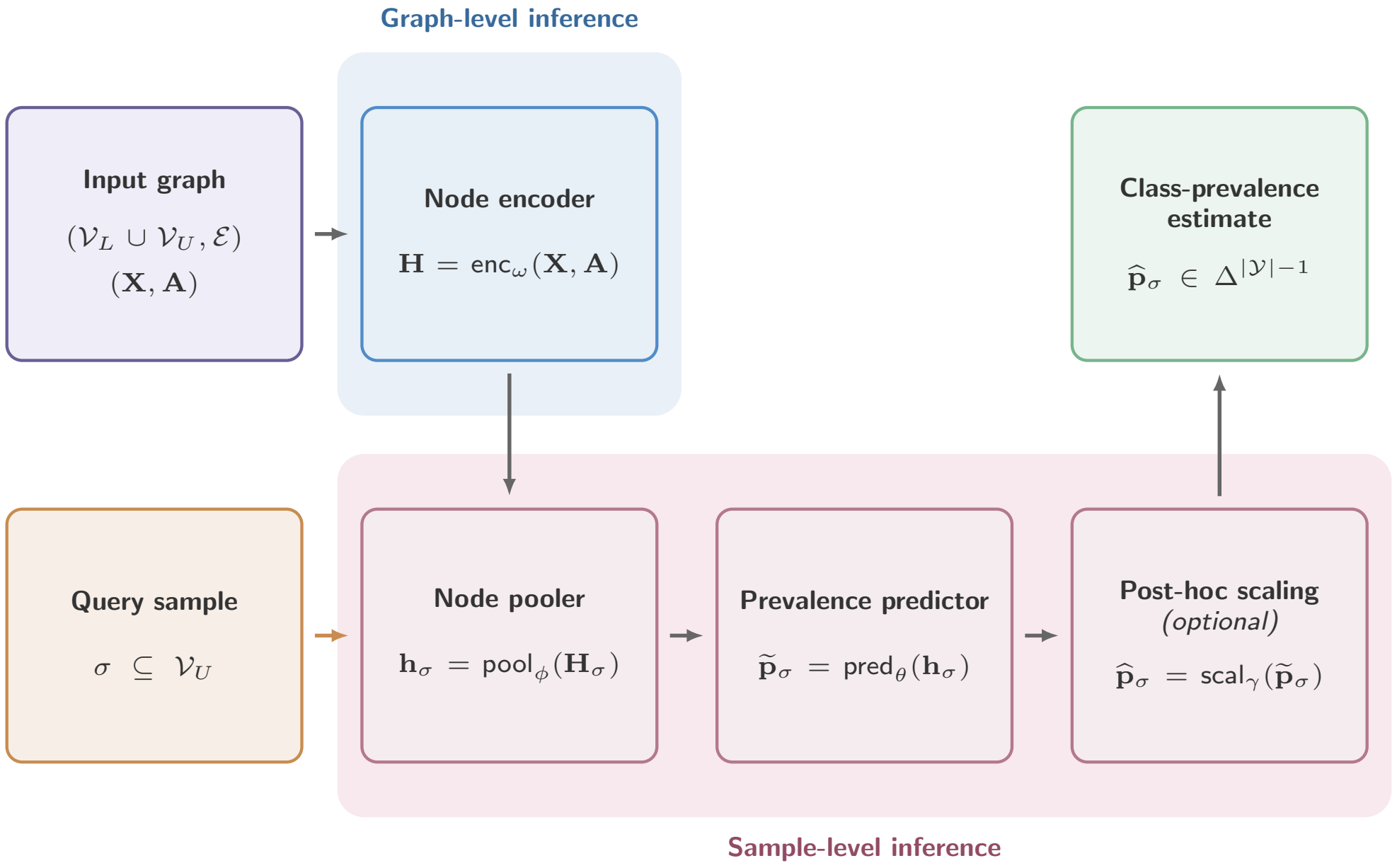
\begin{figure}[t]
  \centering
  \resizebox{\linewidth}{!}{

\begin{tikzpicture}[
  font=\sffamily\footnotesize,
  framework box/.style={
    rectangle,
    rounded corners=4pt,
    line width=0.8pt,
    minimum width=.205\linewidth,
    text width=.18\linewidth,
    minimum height=66pt,
    inner xsep=5pt,
    inner ysep=5pt,
    align=center,
    anchor=center,
    text={rgb,255:red,51;green,51;blue,51},
    fill opacity=1,
    text opacity=1
  },
  flow arrow/.style={
    draw={rgb,255:red,102;green,102;blue,102},
    line width=1pt,
    -{Latex[length=4.5pt,width=3.5pt]},
    shorten <=3pt,
    shorten >=3pt
  },
  query arrow/.style={
    draw={rgb,255:red,200;green,140;blue,80},
    line width=1pt,
    -{Latex[length=4.5pt,width=3.5pt]},
    shorten <=3pt,
    shorten >=3pt
  },
  group label/.style={
    inner sep=0pt,
    font=\sffamily\bfseries\footnotesize
  }
]

\node[framework box,
  draw={rgb,255:red,105;green,95;blue,150},
  fill={rgb,255:red,239;green,237;blue,247}
] (graph) {
  \textbf{Input graph}\\[7pt]
  $(\mathcal V_L\cup\mathcal V_U,\mathcal E)$\\[4pt]
  $(\mathbf X,\mathbf A)$
};

\node[framework box,right=.04\linewidth of graph,
  draw={rgb,255:red,80;green,140;blue,200},
  fill={rgb,255:red,231;green,239;blue,247}
] (encoder) {
  \textbf{Node encoder}\\[9pt]
  $\mathbf H=\component{enc}_{\omega}(\mathbf X,\mathbf A)$
};

\node[framework box,below=38pt of encoder,
  draw={rgb,255:red,180;green,120;blue,140},
  fill={rgb,255:red,245;green,236;blue,239}
] (pooler) {
  \textbf{Node pooler}\\[9pt]
  $\mathbf h_{\sigma}
    =\component{pool}_{\phi}(\mathbf H_{\sigma})$
};

\node[framework box,right=.04\linewidth of pooler,
  draw={rgb,255:red,180;green,120;blue,140},
  fill={rgb,255:red,245;green,236;blue,239}
] (predictor) {
  \textbf{Prevalence predictor}\\[9pt]
  $\widetilde{\mathbf p}_{\sigma}
    =\component{pred}_{\theta}(\mathbf h_{\sigma})$
};

\node[framework box,right=.04\linewidth of predictor,
  draw={rgb,255:red,180;green,120;blue,140},
  fill={rgb,255:red,245;green,236;blue,239}
] (refiner) {
  \textbf{Post-hoc scaling}\\
  \textit{(optional)}\\[6pt]
  $\widehat{\mathbf p}_{\sigma}
    =\component{scal}_{\gamma}(\widetilde{\mathbf p}_{\sigma})$
};

\node[framework box,left=.04\linewidth of pooler,
  draw={rgb,255:red,200;green,140;blue,80},
  fill={rgb,255:red,247;green,239;blue,231}
] (sample) {
  \textbf{Query sample}\\[9pt]
  $\sigma\subseteq\mathcal V_U$
};

\node[framework box,above=38pt of refiner,
  draw={rgb,255:red,120;green,180;blue,140},
  fill={rgb,255:red,236;green,245;blue,239}
] (output) {
  \textbf{Class-prevalence}\\
  \textbf{estimate}\\[7pt]
  $\widehat{\mathbf p}_{\sigma}
    \in\Delta^{|\mathcal Y|-1}$
};

\begin{scope}[on background layer]
  \node[
    rectangle,
    rounded corners=7pt,
    draw=none,
    fill={rgb,255:red,233;green,240;blue,247},
    fit=(encoder),
    inner xsep=6pt,
    inner ysep=14pt,
    outer sep=0pt
  ] (graphphase) {};

  \node[
    rectangle,
    rounded corners=7pt,
    draw=none,
    fill={rgb,255:red,247;green,233;blue,237},
    fit=(pooler)(predictor)(refiner),
    inner xsep=6pt,
    inner ysep=14pt,
    outer sep=0pt
  ] (samplephase) {};
\end{scope}

\node[group label,
  anchor=south,
  text={rgb,255:red,55;green,105;blue,155}
] at ([yshift=5pt]graphphase.north) {
  Graph-level inference
};

\node[group label,
  anchor=north,
  text={rgb,255:red,145;green,75;blue,100}
] (samplephaselabel) at ([yshift=-5pt]samplephase.south) {
  Sample-level inference
};

\draw[flow arrow]
  (graph.east) -- (encoder.west);

\draw[flow arrow]
  (encoder.south) -- (pooler.north);

\draw[query arrow]
  (sample.east) -- (pooler.west);

\draw[flow arrow]
  (pooler.east) -- (predictor.west);

\draw[flow arrow]
  (predictor.east) -- (refiner.west);

\draw[flow arrow]
  (refiner.north) -- (output.south);

\path ([yshift=-18pt]samplephaselabel.south)
  node[inner sep=0pt,outer sep=0pt] {};

\end{tikzpicture}}
 \caption{Inference with GSQ, our proposed graph node symmetric quantification framework. The encoder computes node embeddings, after which, given a query sample $\sigma \subseteq \mathcal{V}_U$, the embeddings of the nodes contained in $\sigma$ are pooled into a sample representation and mapped directly to a class-prevalence estimate, which may optionally undergo post-hoc scaling. The shaded regions distinguish graph-level from sample-level inference.
 }
 \label{fig:symmetric}
\end{figure}
As illustrated in Figure~\ref{fig:symmetric}, class prevalence value
predictions are obtained by composing a \textit{node encoder}
$\component{enc}_{\omega}$, a permutation-invariant \textit{pooler}
$\component{pool}_{\phi}$, a \textit{prevalence predictor}
$\component{pred}_{\theta}$, and a \textit{post-hoc scaling}
$\component{scal}_\gamma$ in the following way
$$
\mathbf{H}=\component{enc}_{\omega}(\mathbf{X},\mathbf{A}) \qquad
\mathbf{h}_\sigma=\component{pool}_{\phi}(\mathbf{H}_\sigma) \qquad
\widetilde{\mathbf{p}}_\sigma=\component{pred}_{\theta}(\mathbf{h}_\sigma)
\qquad \widehat{\mathbf{p}}_{\sigma} =
\component{scal}_\gamma(\widetilde{\mathbf{p}}_{\sigma})
$$
where $\mathbf{H}_\sigma$ contains the embeddings of the nodes in a
given node sample $\sigma$, $\mathbf{h}_\sigma$ is the representation
of the sample,
$\widetilde{\mathbf{p}}_\sigma\in\Delta^{|\mathcal{Y}|-1}$ is a raw
estimate of the class prevalence values in $\sigma$, and
$\widehat{\mathbf{p}}_\sigma\in\Delta^{|\mathcal{Y}|-1}$ is the
(possibly scaled) prevalence vector returned as output.

The node encoder $\component{enc}_{\omega}$ serves the purpose of
generating representations for the nodes in both training and test
samples. Indeed, note that, while an SQ method does not need
\textit{labels} at the individual level for its training data, it
still needs \textit{feature representations} at the individual level,
for both training samples and test samples.

The GSQ framework does not impose a fixed distributional invariance
between the labelled and query populations. Instead, it learns from
labelled node samples, extracted via one of the protocols mentioned in
\S\ref{sec:datasetshiftsimulation}, representative of the query
conditions under which it will be applied.
GSQ is independent of the particular protocol used to construct these
samples, although its effectiveness will ultimately depend on the
labelled samples being representative of the intended query
distribution.

GSQ is also flexible enough to accommodate different encoders, pooling
operators, and prevalence predictors.
We describe these alternatives below, before presenting the training
and evaluation procedure in \S\ref{sec:training-evaluation}. Numerical
settings and experimental configurations are specified in
\S\ref{sec:experiments}.


\subsection{Node encoders}\label{sec:node-encoders}

The role of the node encoder $\component{enc}_\omega$ is to map an
observed graph $(\mathbf{X},\mathbf{A})$ to a matrix of node
embeddings $\mathbf{H}$. We consider an MLP that only operates on node
features as a graph-agnostic baseline. As topology-aware encoders, we
consider two representatives of graph neural networks \citep{Wu2021}
with different inductive biases: GCN, a graph convolutional network
which presents a subgraph motif-oriented bias; GESN, a recursive graph
neural network which presents a node centrality-oriented bias
\citep{Fontanesi2025}.

\subsubsection{Multi-layer perceptron (MLP)}
The MLP computes node representations through $L$ parametrised
non-linear layers
$$
\mathbf{H}^{(\ell)} = \operatorname{ReLU}\left(
  \mathbf{H}^{(\ell-1)}\mathbf{W}^{(\ell)} +\mathbf{b}^{(\ell)}
\right) \qquad \ell=1,\ldots,L
$$
where $\mathbf{H}^{(0)}
=\operatorname{ReLU}(\mathbf{X}\mathbf{W}_{\mathrm{in}}
+\mathbf{b}_{\mathrm{in}})$ and $\mathbf{H}=\mathbf{H}^{(L)}$. Its
trainable parameters are
$$
\omega= \{\mathbf{W}_{\mathrm{in}},\mathbf{b}_{\mathrm{in}}\}
\cup\{\mathbf{W}^{(\ell)}\}_{\ell=1}^{L}
\cup\{\mathbf{b}^{(\ell)}\}_{\ell=1}^{L}
$$
Since each node is processed independently of its neighbours, the MLP
encoder provides a baseline
that enables assessing whether graph topology contributes to
prevalence estimation or not~\citep{errica2020fair}.

\subsubsection{Graph convolutional network (GCN)}
The GCN incorporates topology by combining each node's representation
with those of its neighbours~\citep{Micheli-NN4G,kipf2017gcn}, as
$$
\mathbf{H}^{(\ell)} = \operatorname{ReLU}\left(
  \widetilde{\mathbf{D}}^{-1/2} \widetilde{\mathbf{A}}
  \widetilde{\mathbf{D}}^{-1/2}
  \mathbf{H}^{(\ell-1)}\mathbf{W}^{(\ell)} +\mathbf{b}^{(\ell)}
\right) \qquad \ell=1,\ldots,L
$$
where $\widetilde{\mathbf{A}}=\mathbf{A}+\mathbf{I}$ includes
self-loops and $\widetilde{\mathbf{D}}$ is the diagonal degree matrix
of $\widetilde{\mathbf{A}}$, with
$\widetilde{D}_{ii}=\sum_j\widetilde{A}_{ij}$. Symmetric degree
normalisation weights the contributions of adjacent nodes according to
their degrees, and the self-loops retain each node's own
representation during aggregation. Applying $L$ such layers produces
embeddings that incorporate information from neighbourhoods up to $L$
hops away. The initial representation, final embedding matrix, and
parameter set $\omega$ are defined as for the MLP.

\subsubsection{Graph Echo State Network (GESN)}
GESN~\citep{Gallicchio2010ijcnn, Gallicchio2020gesn,
Micheli2023neurocomp} is a reservoir computing approach which encodes
the graph through a recurrent dynamical system whose parameters remain
fixed, i.e.,
$$
\mathbf{H}^{(t)} = \tanh\left( \mathbf{X}\mathbf{W}_{\mathrm{in}}
  +\mathbf{A}\mathbf{H}^{(t-1)}\mathbf{W}_{\mathrm{rec}} \right)
\qquad t=1,\ldots,T
$$
where $\mathbf{H}^{(0)}=\mathbf{0}$ and the final state
$\mathbf{H}=\mathbf{H}^{(T)}$ is supplied to the pooler. The input
matrix projects node features, while the recurrent matrix transforms
states aggregated from adjacent nodes. Both belong to the reservoir
parameter set
$$
\omega=\{\mathbf{W}_{\mathrm{in}},\mathbf{W}_{\mathrm{rec}}\}
$$
which is randomly initialised and kept frozen during training. The
input weights are sampled uniformly from
$[-\alpha_{\mathrm{in}},\alpha_{\mathrm{in}}]$, and the recurrent
matrix is rescaled so that
$$
\rho(\mathbf{W}_{\mathrm{rec}}) =
\frac{\alpha_{\mathrm{rec}}}{\|\mathbf{A}\|_2}
$$
where $\rho(\cdot)$ denotes the spectral radius. The hyperparameters
$\alpha_{\mathrm{in}}$ and $\alpha_{\mathrm{rec}}$ therefore control
the scales of the feature input and recurrent propagation,
respectively, and are tuned via model selection. Unlike MLP and GCN,
GESN supplies graph-dependent representations without adapting the
encoder to the prevalence targets. This model has been proven
particularly reliable in more challenging node classification tasks
with low homophily \citep{Micheli2023neurocomp}, and was adopted as
the graph encoder backbone in our aggregative node quantification
framework (see \S\ref{sec:xnq}).

\subsection{Node poolers}
\label{sec:node-poolers}

The role of the pooler $\component{pool}_{\phi}$ is to convert the
embeddings of the individual nodes in $\sigma$ into a representation
$\mathbf{h}_\sigma$ of the entire sample $\sigma$ whose dimension is
fixed, irrespective of the size and topology of $\sigma$. As possible
poolers we consider mean pooling, DeepSets~\citep{zaheer2017deep}, and
GMNet~\citep{Perez-Mon:2025jt}, which differ in the distributional
information they retain from the embeddings.

\subsubsection{Mean}
Mean pooling computes the average evidence expressed by the node
embeddings, i.e.,
$$
\mathbf{h}_\sigma=\frac{1}{|\sigma|}\sum_{v\in \sigma}\mathbf{h}_v
$$
It is a parameter-free operator, for which $\phi=\varnothing$. Its
limitation is that samples with different class proportions can have
the same mean embedding, in which case the prevalence predictor
receives identical representations for different targets.

\subsubsection{DeepSets}
DeepSets~\citep{zaheer2017deep} applies learned transformations before
and after averaging, i.e.,
$$
\mathbf{h}_\sigma= \varepsilon\left( \frac{1}{|\sigma|}\sum_{v\in
  \sigma}\varphi(\mathbf{h}_v) \right)
$$
where
\begin{align*}
  \varphi(\mathbf{z})
  &=
    \mathbf{W}_{\varphi}^{(2)}
    \operatorname{ReLU}\left(
    \mathbf{W}_{\varphi}^{(1)}\mathbf{z}
    +\mathbf{b}_{\varphi}^{(1)}
    \right)
    +\mathbf{b}_{\varphi}^{(2)}\\
  \varepsilon(\mathbf{z})
  &=
    \mathbf{W}_{\varepsilon}^{(2)}
    \operatorname{ReLU}\left(
    \mathbf{W}_{\varepsilon}^{(1)}\mathbf{z}
    +\mathbf{b}_{\varepsilon}^{(1)}
    \right)
    +\mathbf{b}_{\varepsilon}^{(2)}
\end{align*}
The transformation $\varphi$ acts independently on each node
embedding, while $\varepsilon$ maps the averaged features to the
sample representation. The set of learnable parameters is
$$
\phi= \{\mathbf{W}_{\varphi}^{(1)},\mathbf{W}_{\varphi}^{(2)},
\mathbf{b}_{\varphi}^{(1)},\mathbf{b}_{\varphi}^{(2)}\} \cup
\{\mathbf{W}_{\varepsilon}^{(1)},\mathbf{W}_{\varepsilon}^{(2)},
\mathbf{b}_{\varepsilon}^{(1)},\mathbf{b}_{\varepsilon}^{(2)}\}
$$
Averaging preserves permutation invariance, but, unlike direct mean
pooling, it operates on learned non-linear features. Therefore,
DeepSets can be in principle more expressive in terms of the set
embeddings it can learn.

\subsubsection{GMNet}
GMNet~\citep{Perez-Mon:2025jt} characterises the distribution of node
embeddings across multiple learned latent spaces. For each view
$j\in\{1,\ldots,J\}$, an independently parametrised network computes
$$
\eta_j(\mathbf{h}_v) = \operatorname{sigmoid}\left( \mathbf{W}_j^{(2)}
  \operatorname{LeakyReLU}\left(
    \mathbf{W}_j^{(1)}\mathbf{h}_v+\mathbf{b}_j^{(1)} \right)
  +\mathbf{b}_j^{(2)} \right)
$$
The projected embeddings are evaluated under $K$ learned Gaussian
densities, whose responses are averaged over the sample, concatenated
across views and components, and linearly projected, i.e.,
$$
\mathbf{h}_\sigma = \mathbf{W}_{\mathrm{pool}} \left[
  \frac{1}{|\sigma|} \sum_{v\in \sigma} \mathcal{N}\left(
    \eta_j(\mathbf{h}_v);
    \boldsymbol{\mu}_{jk},\boldsymbol{\Sigma}_{jk} \right)
\right]_{j=1,k=1}^{J,K} +\mathbf{b}_{\mathrm{pool}}
$$
Here, $\boldsymbol{\mu}_{jk}$ and $\boldsymbol{\Sigma}_{jk}$ are the
mean and positive-definite covariance matrices of component $k$ in
view $j$, and the brackets denote concatenation of the $JK$ scalar
responses. The trainable parameters are
$$
\phi= \{\mathbf{W}_j^{(1)},\mathbf{b}_j^{(1)},
\mathbf{W}_j^{(2)},\mathbf{b}_j^{(2)}\}_{j=1}^{J} \cup
\{\boldsymbol{\mu}_{jk},\boldsymbol{\Sigma}_{jk}\}_{j=1,k=1}^{J,K}
\cup \{\mathbf{W}_{\mathrm{pool}},\mathbf{b}_{\mathrm{pool}}\}
$$
The representations computed by GMNet are permutation-invariant
because the responses are averaged over nodes. The averages describe
how embeddings are distributed across regions of the learned spaces,
allowing GMNet to retain information about multiple modes
that may be lost by a simple mean operator.


\subsection{Prevalence predictors}
\label{sec:prevalence-predictors}

The role of the prevalence predictor $\component{pred}_\theta$ is to
map the sample representation $\mathbf{h}_\sigma$ to a vector
$\widetilde{\mathbf{p}}_\sigma \in \Delta^{|\mathcal{Y}|-1}$
representing a raw estimate of the class prevalence values. As
possible prevalence predictors we consider a softmax head trained with
mean squared error and a Dirichlet head trained with regularised
negative log-likelihood. Throughout this section, we denote by
$\mathcal{D}$ a collection of samples of labelled nodes with known
prevalence vectors.

\subsubsection{Softmax head}
The softmax head applies an affine transformation followed by softmax,
i.e.,
$$
\widetilde{\mathbf{p}}_\sigma = \operatorname{softmax}\left(
  \mathbf{W}_{\mathrm{out}}\mathbf{h}_\sigma
  +\mathbf{b}_{\mathrm{out}} \right) \qquad
\theta=\{\mathbf{W}_{\mathrm{out}},\mathbf{b}_{\mathrm{out}}\}
$$
The resulting entries are non-negative and sum to one, allowing the
predictor to be trained directly against the target prevalences
through the Mean Squared Error (MSE) loss function
$$
\mathcal{L}_{\mathrm{MSE}} = \frac{1}{|\mathcal{D}|\,|\mathcal{Y}|}
\sum_{\sigma\in\mathcal{D}}\sum_{y\in\mathcal{Y}} \left(\widetilde
  p_{\sigma}(y)-p_{\sigma}(y)\right)^2
$$

\subsubsection{Dirichlet head}
The Dirichlet head maps $\mathbf{h}_\sigma$ to positive concentration
parameters and uses the corresponding distributional mean as its
estimate,
i.e.,
$$
\boldsymbol{\alpha}_\sigma = \operatorname{softplus}\left(
  \mathbf{W}_{\mathrm{out}}\mathbf{h}_\sigma
  +\mathbf{b}_{\mathrm{out}} \right)+\epsilon\mathbf{1} \qquad
\widetilde p_{\sigma}(y)=\frac{\alpha_{\sigma}(y)}{\alpha_{\sigma,0}}
$$
where $\alpha_{\sigma,0}=\sum_{y\in\mathcal{Y}}\alpha_{\sigma}(y)$,
$\epsilon>0$, and $\mathbf{1}$ is the all-ones vector. The set of
learned parameters is the same as for the softmax head. Since target
vectors may lie on the simplex boundary, and since the Dirichlet
distribution is supported on the \emph{open} simplex, they are
smoothed and renormalised before evaluating the Dirichlet density,
i.e.,
$$
\widetilde p_{\sigma}(y) = \frac{\max\{p_{\sigma}(y),\epsilon\}}
{\sum_{c\in\mathcal{Y}}\max\{p_{\sigma}(c),\epsilon\}}
$$
Training then minimises the regularised negative log-likelihood loss
function
$$
\mathcal{L}_{\mathrm{Dir}} = \frac{1}{|\mathcal{D}|}
\sum_{\sigma\in\mathcal{D}} \left[ -\log\operatorname{Dir}
  \left(\widetilde{\mathbf{p}}_\sigma;\boldsymbol{\alpha}_\sigma\right)
  +\lambda_{\alpha}\left(\log\alpha_{\sigma,0}\right)^2 \right]
$$
where $\lambda_{\alpha}\geq0$ controls regularisation of the total
concentration. Unlike MSE, this objective learns both the mean and
concentration of a distribution over prevalence vectors, which reduces
to a Beta distribution in the binary case.

\subsection{Post-hoc prevalence scaling}
\label{sec:prevalence-scaling}

Even after training, prevalence estimates may be systematically biased
or overly peaked. We therefore consider a post-hoc transformation
$\component{scal}_\gamma$ with learnable parameters $\gamma$, fitted
on a separate scaling set, with the encoder, pooler, and predictor
held fixed. The transformation receives the raw prevalence estimates
$\widetilde{\mathbf{p}}_\sigma$ and returns a final, scaled estimate
$\widehat{\mathbf{p}}_\sigma$. We evaluate three possibilities,
including leaving the raw estimates unchanged, which corresponds to
setting $\widehat{\mathbf{p}}_\sigma = \widetilde{\mathbf{p}}_\sigma$
and $\gamma = \emptyset$.

\subsubsection{Scalar temperature scaling}
This transformation adjusts raw estimates as
$$
\widehat{\mathbf{p}}_\sigma = \operatorname{softmax}\left(
  \log\widetilde{\mathbf{p}}_\sigma\oslash(\tau\mathbf{1}) \right)
$$
where $\oslash$ denotes element-wise division. For strictly positive
predictions, $\tau>1$ moves the estimates towards the uniform
distribution, whereas $\tau<1$ makes them more concentrated; $\tau=1$
recovers the original predictions. This scaling preserves the ordering
of class prevalences within each sample. This transformation has a
single learnable scalar parameter, i.e., $\gamma = \{\tau \}$.

\subsubsection{Vector scaling}
Vector scaling (VS) \citep{Guo:2017hh} is a variant of temperature
scaling that uses class-specific temperature and bias. VS adjusts the
raw estimates as
$$
\widehat{\mathbf{p}}_\sigma = \operatorname{softmax}\left(
  \log\widetilde{\mathbf{p}}_\sigma\oslash\boldsymbol{\tau}
  +\mathbf{b}-\bar b\mathbf{1} \right) \qquad \bar
b=\frac{1}{|\mathcal{Y}|}\sum_{y\in\mathcal{Y}}b_y
$$
where $\tau_y>0$, $\mathbf{b}$ is a bias term, and $b_y$ is its entry
corresponding to class $y$. Subtracting $\bar b$ centres the biases
without changing the softmax output. This transformation can alter the
ordering of predicted class prevalences, while preserving
non-negativity and unit sum. The learnable parameters of this
transformation are $\gamma = \{ \boldsymbol{\tau}, {\mathbf{b}}\}$.


\subsection{Training and evaluation}
\label{sec:training-evaluation}

In the GSQ framework, the pipeline is trained through sample-level
prevalence supervision, with the selected components determining which
parameters are optimised and the distribution of training samples
determining the query conditions learned by the model. The same
procedure therefore applies to all encoder, pooler, and predictor
combinations.
The set of labelled nodes is partitioned into training, validation,
and scaling sets, i.e.,
$$
\mathcal{V}_L = \mathcal{V}_{\mathrm{train}} \mathbin{\dot\cup}
\mathcal{V}_{\mathrm{val}} \mathbin{\dot\cup}
\mathcal{V}_{\mathrm{scal}}
$$
where $\dot\cup$ denotes disjoint union. Training labels provide the
prevalence targets used to optimise the model, validation labels
support model selection, and scaling labels are reserved for the
post-hoc adjustment. The encoder has access to the complete observed
graph $(\mathbf{X},\mathbf{A})$ throughout, but only labels within the
relevant subset of $\mathcal{V}_L$ are used to construct its samples
and compute their target prevalences.

\subsubsection{Training}
At epoch $e$, a collection $\mathcal{D}^{(e)}=\{\sigma_b\}_{b=1}^{B}$
is constructed from $\mathcal{V}_{\mathrm{train}}$, where $B$ is the
number of samples and each $\sigma_b$ is paired with its true
prevalence vector, calculated according to
Equation~\ref{eq:class-prevalence}. Repeated sampling exposes the
model to different combinations of labelled nodes, although overlap
between samples means that these do not constitute independent
labelled datasets.
For each sample, the pooler aggregates its node embeddings and the
predictor computes a class prevalence estimate. The objective averages
the corresponding per-sample contributions
$$
\mathcal{L}^{(e)}(\omega,\phi,\theta) = \frac{1}{B}
\sum_{\sigma\in\mathcal{D}^{(e)}}
L(\mathbf{\widetilde{p}}_\sigma,\mathbf{p}_\sigma)
$$
where $L$ is the per-sample contribution to either
$\mathcal{L}_{\mathrm{MSE}}$ or $\mathcal{L}_{\mathrm{Dir}}$. Notice
that the objective involves the parameters of the encoder ($\omega$),
those of the pooler ($\phi$), and those of the prevalence predictor
($\theta$), i.e., the three components are not optimized separately,
but jointly. Notice also that we slightly abuse notation in the
Dirichlet case: the loss depends on the full concentration vector
$\boldsymbol{\alpha}_\sigma$, whereas $\widetilde{\mathbf p}_\sigma$
denotes its normalised mean, used as the prevalence estimate.

Samples receive equal weight in the loss, including the undersized
samples that may be produced by the structural protocols.
Gradients propagate through all trainable components, while GESN
encoders and parameter-free poolers remain unchanged.

\subsubsection{Validation, scaling, and model selection}

During validation we use a fixed collection of samples generated from
$\mathcal{V}_{\mathrm{val}}$, so that we compare candidate models on
the same populations. We retain for the subsequent scaling step the
model state corresponding to the lowest mean validation AE. During
scaling, we hold the selected encoder, pooler, and predictor fixed,
while we fit the scalar or vector parameters $\gamma$ separately on
the same samples from $\mathcal{V}_{\mathrm{scal}}$. The objective we
minimise is the cross-entropy
between target and scaled prevalence vectors, regularised towards unit
temperatures and zero-centred biases. Here, we prefer cross-entropy to
other training objectives (e.g., AE or MSE) because it is smoother
than AE, and because it tends to give larger gradients at the simplex
boundaries than MSE.

We fit the two scaling alternatives independently, and we compare them
with each other and with the unchanged predictions (no scaling) on the
validation samples. We subsequently rank component and hyperparameter
configurations by the validation error of their selected scaling
alternative. Throughout this stage, test labels remain obviously
excluded from both fitting the scaler and model selection.

\subsubsection{Evaluation}

Evaluation uses samples drawn from the held-out nodes $\mathcal{V}_U$,
with the selected model and scaling step fixed. For each sample, the
encoder--pooler--predictor mapping produces a prevalence estimate,
which we adjust in the scaling phase (if selected). We measure
performance with mean AE over the test samples, using their true class
prevalence values as targets. Differently from training, we do not not
apply smoothing prior to computing the error.

The reported performance is conditional on the distribution of query
samples used for evaluation. Training and evaluating under
corresponding query conditions assesses generalisation to held-out
nodes under the shift setting of interest, rather than robustness to
arbitrary dataset shifts. Test labels are available only to the
evaluator for computing sample prevalences and quantification error,
and remain unavailable to the predictor.



\section{Experiments}\label{sec:experiments}

We here describe the experiments we have run in order to validate
GSQ. As mentioned in the introduction, we make available all the code
and datasets that are necessary to replicate our results (see
Footnote~\ref{foot:code}).


\subsection{Datasets}
\label{sec:datasets}

We evaluate the GSQ framework on four datasets, namely, Cora-2,
Toloker, Cora, and Amazon-ratings. These cover binary and multi-class
quantification on graphs with different levels of label homophily,
connectivity, and class imbalance. Table~\ref{tab:datasets} summarises
their characteristics. The datasets are the following:

\begin{table}[ht]
  \centering \setlength{\tabcolsep}{7pt}
  \renewcommand{\arraystretch}{1.15}
  \caption{Dataset characteristics. Edges are unique undirected pairs,
  $\bar d$ is the average degree, and $p_{\min}$ and $p_{\max}$ are
  the smallest and largest full-graph class prevalences. Homophily is
  computed using the labels of each task. Cora-2 and Cora share the
  same graph and features.}
  \label{tab:datasets}
  \begin{tabular}{@{}lrrrr@{}}
    \toprule
    \multirow{3}{*}{\textbf{Statistics}}& \multicolumn{2}{c}{\textbf{Binary}} & \multicolumn{2}{c}{\textbf{multi-class}}\\
    \cmidrule(l){2-3} \cmidrule(l){4-5}
                                        & \textbf{Cora-2}
                                                                              & \textbf{Toloker}
                                        & \textbf{Cora}
                                                                              & \textbf{Amazon} \\
    \midrule
    \# of Nodes
                                        & $2{,}708$ & $11{,}758$ & $2{,}708$ & $24{,}492$ \\
    \# of Edges
                                        & $5{,}278$ & $519{,}000$ & $5{,}278$ & $93{,}050$ \\
    \# of Features
                                        & $1{,}433$ & $10$ & $1{,}433$ & $300$ \\
    \# of Classes $|\mathcal{Y}|$
                                        & $2$ & $2$ & $7$ & $5$ \\
    \midrule
    Average degree $\bar d$
                                        & $3.90$ & $88.28$ & $3.90$ & $7.60$ \\
    $p_{\min}$ (\%)
                                        & $15.44$ & $21.82$ & $6.65$ & $4.33$ \\
    $p_{\max}$ (\%)
                                        & $84.56$ & $78.18$ & $30.21$ & $36.79$ \\
    \addlinespace[4pt]
    Edge homophily $h$
                                        & $0.967$ & $0.595$ & $0.810$ & $0.380$ \\
    Adjusted homophily $h_{\mathrm{adj}}$
                                        & $0.886$ & $0.093$ & $0.771$ & $0.140$ \\
    \bottomrule
  \end{tabular}
\end{table}

\begin{itemize}

\item \textbf{Cora} \citep{yang2016cora} is a citation network of
  scientific publications, where nodes represent papers, edges
  represent citations, and each paper is described by a binary
  bag-of-words vector with $1{,}433$ features. We use the Planetoid
  version of the dataset, whose labels distinguish seven research
  topics. The multi-class task estimates the proportions of these
  topics within each queried sample.

\item \textbf{Cora-2} uses the same graph and features as Cora, with
  the original class indexed by $2$ designated as positive and the
  remaining classes merged into the negative class. The positive class
  contains $418$ nodes, corresponding to $15.44\%$ of the graph. This
  construction allows binary and multi-class quantification to be
  compared while holding the graph and attributes fixed. Since merging
  classes also changes label agreement across edges, homophily is
  computed separately for each task.

\item \textbf{Toloker} \citep{Likhobaba2023toloker} is a network of
  workers on the Toloka crowdsourcing platform, drawn from thirteen
  projects. Edges connect workers who have completed a common task,
  while ten node features describe their profile and task
  performance. The binary label indicates whether a worker was banned
  in one of the projects; this class accounts for $21.82\%$ of the
  nodes. With an average degree of $88.28$, Toloker provides a
  substantially denser graph than the other datasets.

\item Amazon-ratings (hereafter: \textbf{Amazon})
  \citep{platonov2023questions-tolokers} is a product co-purchasing
  network whose edges connect products frequently bought together, and
  node labels are product ratings. Node features are $300$-dimensional
  averages of FastText embeddings of words in product descriptions,
  and labels group average review ratings into five categories. The
  released graph contains the largest connected component of the
  original network's $5$-core.
  Coherently with most literature, we treat the 5 rating categories
  (ranging from 1 to 5) as nominal classes and estimate their
  proportions within each sample.

\end{itemize}

\subsubsection{Graph characteristics}
All graphs are treated as undirected, and the edge counts in
Table~\ref{tab:datasets} refer to unique node pairs, excluding
possible self-loops introduced by an encoder. To describe class
imbalance, we report the smallest and largest full-graph class
prevalences. We also report edge homophily (i.e., the proportion of
edges whose endpoints have the same class label),
defined as
$$
h= \frac{ \sum_{\{u,v\}\in\mathcal{E}} \mathbb{I}[y_u=y_v] }{
|\mathcal{E}| }
$$
where $\mathcal{E}$ contains each undirected edge once. Because this
measure depends on the class distribution and its relation to node
degree, we additionally report adjusted homophily:
$$
h_{\mathrm{adj}} = \frac{h-\sum_{y\in\mathcal{Y}}q_y^2}
{1-\sum_{y\in\mathcal{Y}}q_y^2} \qquad q_y=
\frac{\sum_{v:y_v=y}\deg(v)} {\sum_{v\in\mathcal{V}}\deg(v)}
$$
The adjustment measures agreement relative to the baseline determined
by the class proportions at edge endpoints. Both Cora tasks are
strongly homophilic, whereas Toloker and Amazon-ratings have
substantially lower adjusted homophily and are used as heterophily
benchmarks. These differences provide contrasting evaluation settings,
although comparisons across datasets cannot isolate the effect of
homophily from that of graph size, features, or class imbalance.

\subsubsection{Data partitioning}
For each dataset, we partition the nodes into disjoint training
($\sim$40\% of the nodes), scaling ($\sim$20\%), validation
($\sim$20\%), and test ($\sim$20\%) sets that collectively cover the
graph. In the notation of \S\ref{sec:training-evaluation}, the first
three sets form $\mathcal{V}_L=\mathcal{V}_{\mathrm{train}}\cup
\mathcal{V}_{\mathrm{val}}\cup\mathcal{V}_{\mathrm{scal}}$, while the
test set is $\mathcal{V}_U$, whose labels are withheld from the
learner and used only for evaluation. The partitioning is performed in
a stratified way, so that each of the four sets is characterised by
approximately the same node label distribution. The complete topology
and node features remain available irrespectively of the set, while
node labels are only accessible within the respective set. Notice
that, according to common splitting practice, distant nodes may belong
to the same set, meaning that the partitioning is performed without
enforcing any proximity constraint.
For each dataset we actually generate five different such partitions,
by rotating the nodes in a five-fold cross-validation fashion, so that
each node (and each sample generated by the protocol used in the given
experiment) appears twice in training and once in the other three
roles. The two Cora tasks use identical node partitions.


\subsection{Experimental protocols}
\label{sec:experimental-protocols}
\noindent We use three sample-generation protocols to simulate varying
amounts of dataset shift. The APP protocol (a standard in
quantification that goes back to~\cite{Forman:2005fk}) generates prior
probability shift, whereas the BFP and RWP protocols (which are
inspired by similar ones proposed by \citet{Damke:2025qf}) produce
structurally covariate shift. Let us remark that these protocols, as
all protocols, each operationalise a specific shift setting for
experimental purposes, and do \textit{not} constitute a definition of
such a setting.



\begin{figure}[t]
  \centering
  \begin{subfigure}[b]{0.33\linewidth}
    \centering \resizebox{3.8cm}{!}{\begin{tikzpicture}[x=1mm,y=1mm]
  \definecolor{classblue}{HTML}{72A6D3}
  \definecolor{classred}{HTML}{CC6677}
  \definecolor{classgreen}{HTML}{228833}
  \definecolor{edgegray}{HTML}{AEB5BF}
  \definecolor{ink}{HTML}{111827}

  \path[use as bounding box] (0,0) rectangle (86,76);

  \begin{scope}[
    edge/.style={draw=edgegray,line width=0.55pt,line cap=round},
    node/.style={circle,minimum size=4.4mm,inner sep=0pt,draw=white,line width=0.6pt},
    member/.style={circle,minimum size=6.6mm,inner sep=0pt,draw=ink,line width=0.75pt}
  ]
    \draw[edge] (10,61)--(25,70)--(42,60)--(59,70)--(75,57);
    \draw[edge] (10,61)--(16,43)--(31,47)--(42,60);
    \draw[edge] (25,70)--(31,47)--(50,40)--(59,70);
    \draw[edge] (75,57)--(70,39)--(50,40);
    \draw[edge] (16,43)--(8,24)--(25,21)--(31,47);
    \draw[edge] (31,47)--(25,21)--(45,16)--(50,40);
    \draw[edge] (50,40)--(45,16)--(66,21)--(70,39);
    \draw[edge] (8,24)--(18,3)--(38,3)--(45,16);
    \draw[edge] (45,16)--(38,3)--(60,2)--(66,21);

    \node[node,fill=classblue]  at (10,61) {};
    \node[node,fill=classred]   at (25,70) {};
    \node[node,fill=classgreen] at (42,60) {};
    \node[node,fill=classblue]  at (59,70) {};
    \node[node,fill=classred]   at (75,57) {};
    \node[node,fill=classgreen] at (16,43) {};
    \node[node,fill=classblue]  at (31,47) {};
    \node[node,fill=classred]   at (50,40) {};
    \node[node,fill=classgreen] at (70,39) {};
    \node[node,fill=classblue]  at (8,24) {};
    \node[node,fill=classred]   at (25,21) {};
    \node[node,fill=classgreen] at (45,16) {};
    \node[node,fill=classblue]  at (66,21) {};
    \node[node,fill=classgreen] at (18,3) {};
    \node[node,fill=classblue]  at (38,3) {};
    \node[node,fill=classred]   at (60,2) {};

    \node[member] at (10,61) {};
    \node[member] at (42,60) {};
    \node[member] at (16,43) {};
    \node[member] at (50,40) {};
    \node[member] at (8,24) {};
    \node[member] at (45,16) {};
    \node[member] at (38,3) {};
    \node[member] at (60,2) {};
  \end{scope}
\end{tikzpicture}}
    \caption{APP}
    \label{fig:shift-pps}
  \end{subfigure}%
  \begin{subfigure}[b]{0.33\linewidth}
    \centering \resizebox{3.8cm}{!}{\begin{tikzpicture}[x=1mm,y=1mm]
  \definecolor{classblue}{HTML}{72A6D3}
  \definecolor{classred}{HTML}{CC6677}
  \definecolor{classgreen}{HTML}{228833}
  \definecolor{edgegray}{HTML}{AEB5BF}
  \definecolor{ink}{HTML}{111827}
  \definecolor{traversal}{HTML}{B7791F}

  \path[use as bounding box] (0,0) rectangle (86,76);

  \begin{scope}[
    edge/.style={draw=edgegray,line width=0.55pt,line cap=round},
    node/.style={circle,minimum size=4.4mm,inner sep=0pt,draw=white,line width=0.6pt},
    member/.style={circle,minimum size=6.6mm,inner sep=0pt,draw=ink,line width=0.75pt},
    traversed/.style={draw=traversal,line width=2.2pt,line cap=round,line join=round}
  ]
    \draw[edge] (10,61)--(25,70)--(42,60)--(59,70)--(75,57);
    \draw[edge] (10,61)--(16,43)--(31,47)--(42,60);
    \draw[edge] (25,70)--(31,47)--(50,40)--(59,70);
    \draw[edge] (75,57)--(70,39)--(50,40);
    \draw[edge] (16,43)--(8,24)--(25,21)--(31,47);
    \draw[edge] (31,47)--(25,21)--(45,16)--(50,40);
    \draw[edge] (50,40)--(45,16)--(66,21)--(70,39);
    \draw[edge] (8,24)--(18,3)--(38,3)--(45,16);
    \draw[edge] (45,16)--(38,3)--(60,2)--(66,21);

    \draw[traversed] (31,47)--(16,43);
    \draw[traversed] (31,47)--(50,40);
    \draw[traversed] (31,47)--(25,21)--(45,16)--(38,3);

    \node[node,fill=classblue]  at (10,61) {};
    \node[node,fill=classred]   at (25,70) {};
    \node[node,fill=classgreen] at (42,60) {};
    \node[node,fill=classblue]  at (59,70) {};
    \node[node,fill=classred]   at (75,57) {};
    \node[node,fill=classgreen] at (16,43) {};
    \node[node,fill=classblue]  at (31,47) {};
    \node[node,fill=classred]   at (50,40) {};
    \node[node,fill=classgreen] at (70,39) {};
    \node[node,fill=classblue]  at (8,24) {};
    \node[node,fill=classred]   at (25,21) {};
    \node[node,fill=classgreen] at (45,16) {};
    \node[node,fill=classblue]  at (66,21) {};
    \node[node,fill=classgreen] at (18,3) {};
    \node[node,fill=classblue]  at (38,3) {};
    \node[node,fill=classred]   at (60,2) {};

    \node[member] at (16,43) {};
    \node[member] at (31,47) {};
    \node[member] at (50,40) {};
    \node[member] at (25,21) {};
    \node[member] at (45,16) {};
    \node[member] at (38,3) {};

    \fill[ink] (31,47) circle[radius=1mm];
  \end{scope}
\end{tikzpicture}}
    \caption{BFP}
    \label{fig:shift-bfs}
  \end{subfigure}%
  \begin{subfigure}[b]{0.33\linewidth}
    \centering \resizebox{3.8cm}{!}{\begin{tikzpicture}[x=1mm,y=1mm]
  \definecolor{classblue}{HTML}{72A6D3}
  \definecolor{classred}{HTML}{CC6677}
  \definecolor{classgreen}{HTML}{228833}
  \definecolor{edgegray}{HTML}{AEB5BF}
  \definecolor{ink}{HTML}{111827}
  \definecolor{walk}{HTML}{76558F}

  \path[use as bounding box] (0,0) rectangle (86,76);

  \begin{scope}[
    edge/.style={draw=edgegray,line width=0.55pt,line cap=round},
    node/.style={circle,minimum size=4.4mm,inner sep=0pt,draw=white,line width=0.6pt},
    member/.style={circle,minimum size=6.6mm,inner sep=0pt,draw=ink,line width=0.75pt}
  ]
    \draw[edge] (10,61)--(25,70)--(42,60)--(59,70)--(75,57);
    \draw[edge] (10,61)--(16,43)--(31,47)--(42,60);
    \draw[edge] (25,70)--(31,47)--(50,40)--(59,70);
    \draw[edge] (75,57)--(70,39)--(50,40);
    \draw[edge] (16,43)--(8,24)--(25,21)--(31,47);
    \draw[edge] (31,47)--(25,21)--(45,16)--(50,40);
    \draw[edge] (50,40)--(45,16)--(66,21)--(70,39);
    \draw[edge] (8,24)--(18,3)--(38,3)--(45,16);
    \draw[edge] (45,16)--(38,3)--(60,2)--(66,21);

    \draw[
      walk,
      line width=2.2pt,
      line cap=round,
      line join=round
    ]
      (25,70)--(42,60)--(31,47)--(50,40)--(45,16)--
      (25,21)--(8,24)--(18,3)--(38,3)--(45,16)--(66,21);

    \node[node,fill=classblue]  at (10,61) {};
    \node[node,fill=classred]   at (25,70) {};
    \node[node,fill=classgreen] at (42,60) {};
    \node[node,fill=classblue]  at (59,70) {};
    \node[node,fill=classred]   at (75,57) {};
    \node[node,fill=classgreen] at (16,43) {};
    \node[node,fill=classblue]  at (31,47) {};
    \node[node,fill=classred]   at (50,40) {};
    \node[node,fill=classgreen] at (70,39) {};
    \node[node,fill=classblue]  at (8,24) {};
    \node[node,fill=classred]   at (25,21) {};
    \node[node,fill=classgreen] at (45,16) {};
    \node[node,fill=classblue]  at (66,21) {};
    \node[node,fill=classgreen] at (18,3) {};
    \node[node,fill=classblue]  at (38,3) {};
    \node[node,fill=classred]   at (60,2) {};

    \node[member] at (25,70) {};
    \node[member] at (42,60) {};
    \node[member] at (31,47) {};
    \node[member] at (50,40) {};
    \node[member] at (45,16) {};
    \node[member] at (25,21) {};
    \node[member] at (8,24) {};
    \node[member] at (18,3) {};
    \node[member] at (38,3) {};
    \node[member] at (66,21) {};

    \fill[ink] (25,70) circle[radius=1mm];

    \begin{scope}[
      direction/.style={
        draw=walk,
        line width=2.2pt,
        line cap=round,
        -{Latex[length=4.6mm,width=3.4mm]}
      }
    ]
      \draw[direction] (30.05,67.03) -- (35.48,63.83);
      \draw[direction] (39.08,56.55) -- (35.01,51.75);
      \draw[direction] (36.75,44.88) -- (42.66,42.70);
      \draw[direction] (48.32,31.92) -- (47.03,25.75);
      \draw[direction] (38.88,17.53) -- (32.77,19.06);
      \draw[direction] (20.44,21.80) -- (14.24,22.90);
      \draw[direction] (11.28,17.11) -- (13.99,11.42);
      \draw[direction] (24,3) -- (30.3,3);
      \draw[direction] (39.60,5.98) -- (42.59,11.53);
      \draw[direction] (51.61,17.57) -- (57.74,19.03);
    \end{scope}
  \end{scope}
\end{tikzpicture}}
    \caption{RWP}
    \label{fig:shift-rw}
  \end{subfigure}
 \caption{The three sampling protocols considered in this study. 
 Node colour denotes class membership, outlined nodes belong to the sample, and the inner black circle identifies a seed node. (a) APP samples eligible nodes with replacement to realise a prescribed class prevalence, independently of graph locality. (b) BFP runs a breadth-first traversal of the complete graph from a seed node (here, up to three hops), but adds only eligible visited nodes to the sample. Consequently, closer non-outlined nodes may be traversed without being included.
 (c) RWP performs a stochastic crawl from a seed and adds each
 eligible visited node at most once. APP therefore isolates controlled
 prevalence shift, whereas BFP and RWP induce prevalence variation
 through structural sampling.}\label{fig:shifts}
\end{figure}
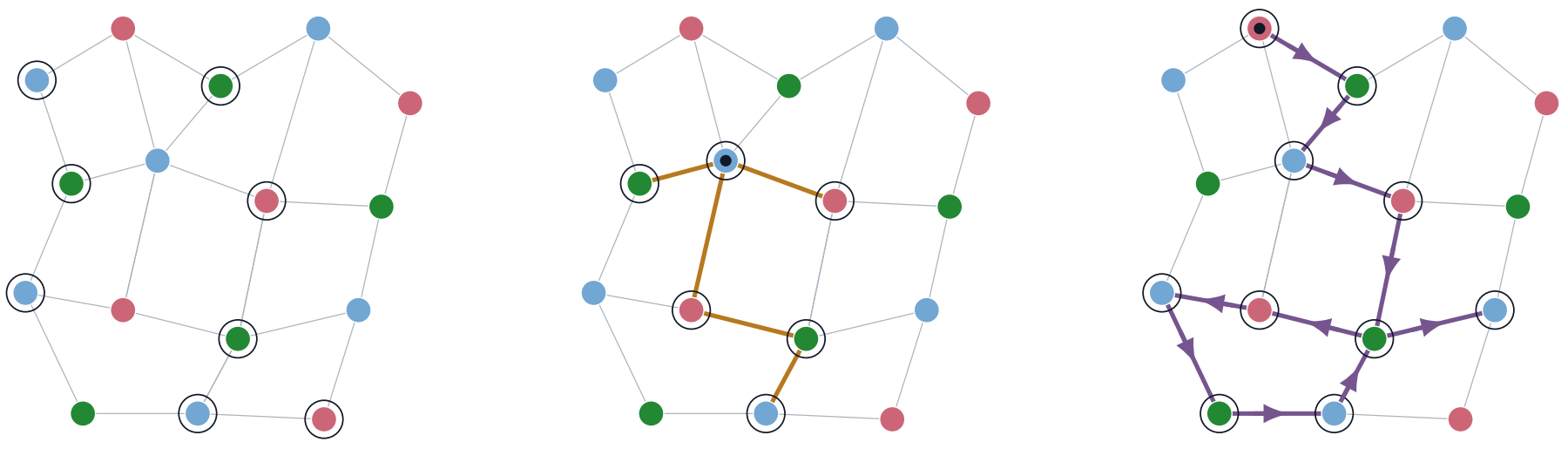

\subsubsection{Simulating prior probability shift via the APP}
The experimental protocol for testing quantification systems that has
been dominant in the literature (the so-called \emph{artificial
prevalence protocol} -- APP) is one that indeed simulates prior
probability shift. In the binary case, this protocol consists of
randomly extracting from the test set a number of test samples,
obtained by sampling with replacement, characterised by class
frequency values for the minority class that lie on a predefined grid
of values (e.g., the 21-point grid \{0.00, 0.05, ..., 0.95, 1.00\}).
These samples are the test samples on which each quantifier will be
tested; therefore, the result of using this protocol is that of
``stress-testing'' the quantifiers, i.e., testing their ability to
correctly guess class frequency values possibly much different from
the ones seen at training time. This experimental protocol clearly
simulates the presence of prior probability shift, since, assuming
that the training set and the test set (from which the test samples
are extracted) are IID, for each test sample it is true that (a)
$P(Y)$ has changed with respect to the training set (since, for each
sample, a specific class frequency from the grid has been imposed),
and (b) $P(X|Y)$ has not changed, since all the data points in all the
samples are from the original test set. The selected nodes may
consequently belong to distant graph regions, as shown in
Figure~\ref{fig:shift-pps}.

The APP is well-defined for the multi-class case too. The only
difference between the binary case and the multi-class case is that in
the latter, due to the fact that using a predefined grid of values
would generate a combinatorial explosion, we sample (as first
suggested by~\citet{Esuli:2022hy}) points on the probability simplex
$\Delta^{|\mathcal{Y}|-1}$ uniformly at random via a robust protocol
such as the Kraemer algorithm. Such points represent the desired
prevalence vectors that each of the generated samples realise via
subsampling.

The protocol simulates realistic situations displaying prior
probability shift; for instance, for a disease prediction task the
training data may have been collected in a pre-pandemic situation
while the unlabelled data may originate from a pandemic scenario, with
both data collections performed uniformly at random.



\subsubsection{Simulating structural covariate shift via the BFP}
The breadth-first protocol (BFP) constructs a sample around a seed
$v_0\in \mathcal{V}'$ by running a standard breadth-first search on
the complete observed graph. The traversal visits the seed's one-hop
neighbours first, followed by nodes at two hops, three hops, and so
forth. A visited node enters $\sigma$ only if it belongs to the set
$\mathcal{V}'$ of nodes eligible for the active node set (e.g., the
training set). Consequently, an eligible node several hops from the
seed may be included even when closer ineligible nodes are not; the
latter may nevertheless serve as intermediate traversal nodes.
%
Figure~\ref{fig:shift-bfs} illustrates a sample generated by this
protocol. Traversal stops when the requested sample size or the
prescribed exploration limit is reached, or when no further nodes are
reachable. Each eligible node is included at most once, and undersized
samples are retained at their realised size rather than padded with
repeated nodes. The resulting prevalence is therefore determined by
the class composition of the selected neighbourhood, rather than
specified in advance.

This protocol simulates a situation that may realistically occur in
graph-based settings, i.e., the training data are from a certain,
well-connected region of the graph while the unlabelled data are from
a different one. For instance, we may have labelled data from towns in
Sweden (where towns are nodes and railway connections are edges) while
data from towns in Norway may be unavailable.


\subsubsection{Simulating structural covariate shift via the RWP}
The \textit{random-walk protocol} (RWP) constructs samples through a
stochastic traversal of the complete observed graph, as illustrated in
Figure~\ref{fig:shift-rw}. Starting from a seed $v_0\in\mathcal{V}'$,
the walker either restarts at the seed or moves to a uniformly
selected neighbour, adding a visited node to $\sigma$ only if it is
eligible and has not already been included. The walk terminates upon
reaching the requested sample size or its transition budget, with
undersized samples retained at their realised size.

The protocol simulates realistic situations displaying structural
covariate shift; for instance, in an online marketplace, labelled
products may have been collected by sampling uniformly at random,
while the products queried at inference time may have been reached by
repeatedly following recommendation links.

\bigskip

Both structural protocols preserve the observed topology during
traversal while restricting sample membership to eligible nodes. The
resulting samples may therefore differ in terms of $P(X)$ and $P(Y)$,
and exhibit either covariate shift or prior probability shift,
depending on the nature of the problem.

Notice that the BFP and RWP protocols are inspired by those proposed
by \citet{Damke:2025qf}, but differ in how they traverse the graph and
construct samples. BFP operates on the complete observed graph rather
than the split-induced graph $G[\mathcal{V}']$, while RWP performs an
explicit walk rather than sampling from a diffusion-based endpoint
distribution. These changes have the effect of preserving intermediate
connections and avoiding weighting sample prevalences through repeated
occurrences of the same node.


\subsection{Aggregative baselines}
\label{sec:aggregative-baselines}
\noindent We compare the proposed GSQ symmetric framework with
aggregative methods under the same node partitions and sampling
protocols. Below, we document the two aggregative baselines taken from
the node quantification literature. All models are selected by
validation quantification error and evaluated on identical test
samples. The distinction is therefore in the use of supervision:
aggregative methods learn individual node predictions before
estimating prevalences, whereas the symmetric methods optimise
sample-level prevalence objectives. The different hyperparameters are
summarised in Table~\ref{tab:aggregative-hyperparameters}, while
shared training settings are instead described in
\S\ref{sec:experimental-setup}.

\subsubsection{eXtreme Network Quantifier (XNQ)}
\label{sec:xnq}
The eXtreme Network Quantifier (XNQ)~\citep{Micheli:2025ez} combines
frozen graph reservoir embeddings with a trained logistic regression
readout and the Saerens--Latinne--Decaestecker (SLD) adjustment. SLD
iteratively updates the estimated class prevalences and adjusts node
posterior probabilities under PPS using a calibration set
($\mathcal{V}_{\mathrm{scal}}$ in our experiments). Since XNQ and the
GESN-based symmetric methods share the same encoder family, their
comparison helps assess the effect of replacing node classification
and aggregative correction with direct prevalence learning.

\subsubsection{Approximated Personalised Propagation of Neural
Predictions (APPNP)}
This baseline combines an Approximated Personalised Propagation of
Neural Predictions (APPNP) probabilistic node classifier
\citep{klicpera2019predict} with KDEy \citep{Moreo:2025lq}. KDEy
represents each class through a class-conditional kernel density
estimate of the classifier probability vectors observed on the
labelled reference nodes $\mathcal{V}_{\mathrm{scal}}$, and estimates
prevalence by matching their mixture to the distribution observed in
the query sample.

For structural sampling protocols, we additionally consider the
structural importance sampling (SIS) correction of
\citet{Damke:2025qf}, which adapts the class-conditional densities to
each query by weighting reference nodes according to their structural
proximity to the sampled nodes. We denote this method by
APPNP-KDEy-SIS and retain ordinary APPNP-KDEy as the zero-weight
alternative. Both methods use raw APPNP probabilities, and SIS is
applied only at quantification time without changing the classifier
training objective. Its interpolation weight is selected according to
validation AE.


\subsection{Experimental setup}
\label{sec:experimental-setup}

We implemented the models using prevalently
PyTorch~\citep{Paszke:2019oq}, PyTorch
Geometric~\citep{torch-geometric2019}, and QuaPy~\citep{Moreo:2021bs},
using the fixed node partitions described above. Each architecture is
selected separately for each dataset, fold, and sampling protocol,
with the same validation and test samples used across methods. The
encoder always observes the complete graph, while the eligible nodes
and available labels are determined by the active partition. The
different hyperparameters per component type are summarised in
Table~\ref{tab:framework-hyperparameters}.

\subsubsection{Component configurations} 
For MLP and GCN, we search over hidden dimensions $\{64,128\}$ and
depths $\{2,3\}$. Both encoders begin with a linear input projection
and ReLU activation; each subsequent layer is followed by layer
normalisation, ReLU, and dropout. The depth parameter counts these
subsequent layers, excluding the input projection.

GESN uses a single frozen reservoir layer, with its dimension and
input and recurrent scales selected from their respective grids. When
paired with DeepSets or GMNet, its embeddings are centred and
projected by PCA, using the paired dimensions $(1024,64)$ and
$(2048,128)$ for the original and projected representations. This
reduction limits the computational cost and number of trainable
parameters introduced by applying learned poolers to high-dimensional
reservoir states. PCA is fitted to all observed node embeddings
without using labels, consistently with the transductive setting, and
remains fixed during training. Mean pooling uses the original
reservoir embeddings. Pooler and predictor settings are otherwise
fixed, so the architectural search concerns the encoder rather than an
additional grid for every component.

The models are not explicitly matched by parameter count, but their
trainable parameter ranges are broadly comparable, with all
configurations remaining below half a million
parameters. Frozen-reservoir methods with mean pooling require
substantially fewer learned parameters, while MLP and GCN have
identical counts at corresponding widths and depths, as do the softmax
and Dirichlet heads.

\subsubsection{Sample generation}
Under PPS, each sample contains $500$ node draws. Binary tasks use APP
with $21$ target prevalences in the grid $\{0.00, 0.05, \dots 0.95,
1.00\}$, while multi-class tasks use Kraemer's algorithm. Thus, per
training epoch, APP generates $B=21\times128$ samples for binary
tasks, and $B=128$ samples for multi-class tasks. Refinement,
validation, and test samples each contain $100$ APP samples; therefore
$B=21 \times100$ for binary tasks and $B=100$ for multi-class tasks.
Sampling within each class follows QuaPy's sampling-with-replacement
procedure, so PPS samples may contain repeated nodes.

BFP and RWP collect distinct eligible nodes, with each sample
containing at most $100$ nodes. BFP uses ten seeds per class and a
maximum traversal depth of ten, while RWP uses $100$ uniformly
selected seeds, a restart probability of $0.1$, and a limit of
$10{,}000$ transitions. Since each seed produces one sample, BFP
generates $B=10|\mathcal{Y}|$ samples and RWP generates $B=100$
samples. These values are used during training, validation, and test
evaluation. Both protocols traverse the complete observed graph and
retain undersized samples without padding when fewer than $100$
eligible nodes can be reached (e.g., when the number of traversable
nodes is less than $100$). During structural training, we draw a
class-stratified subset of the training nodes whose size matches the
validation partition, thereby avoiding a higher density of eligible
nodes solely because the training partition is larger.

Regardless of the training protocol, the $B$ training samples are
regenerated at every epoch, whereas the validation samples remain
fixed.

\subsubsection{Optimisation and selection}
All trainable models use AdamW with learning rate $10^{-3}$ and weight
decay $10^{-5}$, with gradient values clipped to $[-1,1]$. Symmetric
models minimise their sample-level objective, whereas aggregative
classifiers are trained with node-level cross-entropy. Frozen
reservoir representations and their PCA projections are computed once
and reused.

Training runs for at most $500$ epochs, with validation every 10
epochs and a minimum of 20 epochs. The learning rate is halved after
five validation checks without sufficient improvement, down to
$10^{-6}$; early stopping uses a patience of ten validation checks.
Both mechanisms monitor validation AE with an improvement threshold of
$5\times10^{-5}$, and the model state with the lowest validation AE is
retained.

For the post-hoc stage, scalar and vector scaling are fitted on the
scaling samples and compared with the raw predictions using validation
AE. Refinement minimises regularised cross-entropy with L-BFGS,
starting from the identity transformation. XNQ applies the
corresponding scaling alternatives to node probabilities, using node
labels for fitting and downstream quantification AE for selection.

The experiment configuration uses the same random seed across all
experiments and five precomputed CV folds to ensure
reproducibility. Hyperparameters are selected within each fold, and
performance is summarised across folds using mean test AE.

\subsubsection{Computational resources}

The complete experimental search comprises 10,620 model-training runs,
including all hyperparameter candidates, five data folds, four
datasets, and three sampling protocols. Post-hoc scaling alternatives
were fitted within each run and therefore did not require separate
model-training jobs. The experiments were executed on a Linux server
equipped with 256 logical CPU cores and four NVIDIA A100-SXM4 GPUs
with 40\,GB of memory each. The entire hyperparameter search required
approximately 372 allocated GPU-hours and 15,500 allocated
CPU-core-hours, while the final evaluation added approximately 32
CPU-core-hours.

\FloatBarrier


\section{Results and discussion}
\label{sec:results}


\subsection{Comparison with aggregative methods}
\label{sec:comparisonwithaggregative}
\begin{figure}[h!]
  \centering \centering
  \includegraphics[width=\linewidth]{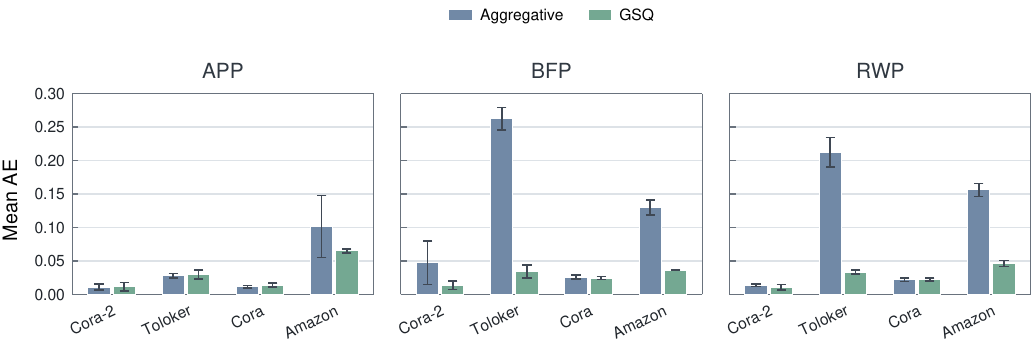}
  \caption{Test performance of the validation-selected aggregative and
  symmetric configurations under the three sampling protocols
  considered in this study. For each dataset, sampling protocol, and
  method family, the configuration with the best mean validation AE
  (lower is better) was selected independently. Bars report its mean
  test AE across the five folds, and error bars indicate one standard
  deviation.}
  \label{fig:comparison}
\end{figure}

\noindent For each dataset and sampling protocol, we independently
selected the aggregative and symmetric configurations with the lowest
mean validation AE. Figure~\ref{fig:comparison} reports the test
performance of the resulting validation-selected configurations; test
results were not used during selection. The results are discussed
separately for PPS, BFP, and RWP because the protocols induce
substantially different dataset shifts.

\subsubsection{APP results}
\label{sec:PPS}

\noindent
Under prior probability shift, XNQ is the validation-selected
aggregative method on all four datasets. It achieves a lower mean test
AE than the selected symmetric configuration on Cora-2, Toloker, and
Cora, although the differences are small and should be treated as
descriptive rather than statistically significant. This agrees with
previous evidence that XNQ performs well under prior probability shift
across graphs with different levels of
homophily~\citep{Micheli:2025ez}.

Amazon is the only exception, where the selected
GESN--GMNet--Dirichlet configuration achieves a lower mean test AE
than XNQ. The performance of XNQ also varies substantially across
folds, as indicated by its larger standard deviation. Overall,
symmetric methods remain competitive under prior probability shift but
do not generally improve over a strong aggregative method designed for
this setting. The complete results are presented in
Appendix~\ref{app:results-pps}; Appendix~\ref{app:pps-diagonal}
further characterizes the binary results using diagonal plots.




\subsubsection{BFP results}
\label{sec:BFP}

Under breadth-first sampling, the validation-selected symmetric
configuration outperforms its aggregative counterpart on every
dataset. The difference is largest on Toloker and Amazon, remains
substantial on Cora-2, and is comparatively small on Cora.

This reversal relative to prior probability shift is associated with
the sampling protocol, since the data splits, selection procedure, and
training seeds are otherwise unchanged. Non-aggregative models are
trained directly on samples generated through breadth-first traversal
and can therefore learn the relationship between the resulting sample
representations and their prevalences. Among the aggregative
alternatives, XNQ does not explicitly account for structural covariate
shift, while KDEy-SIS adapts its class-conditional density estimates
through a predefined structural weighting model. Its performance on
Toloker indicates that this correction does not always capture the
distributional changes induced by breadth-first sampling. The smaller
difference on Cora nevertheless shows that the effect is dataset
dependent and that aggregative methods do not necessarily fail under
every structural sampling condition. The complete results are
presented in Appendix~\ref{app:results-bfs}.



\subsubsection{RWP results}
\label{sec:RWP}

The results under random-walk sampling closely resemble those obtained
under breadth-first sampling. The validation-selected symmetric
configuration achieves a lower mean test AE on Cora-2, Toloker, and
Amazon. The difference is large on Toloker and Amazon and smaller on
Cora-2. On Cora, APPNP-KDEy-SIS and the selected symmetric
configuration obtain mean test AEs of $0.0217$ and $0.0221$,
respectively, and are therefore effectively tied.

Across the two structural protocols, the selected symmetric
configuration achieves a lower mean test AE in seven of the eight
comparisons. Direct training on samples generated through the
corresponding traversal process is therefore generally more effective
than applying an aggregative correction to node-level predictions,
although the Cora result under random-walk sampling shows that the
advantage is not universal. Together with the prior probability shift
results, these findings support a qualified conclusion: aggregative
methods are strongest under controlled prevalence shift, whereas the
GSQ framework is generally preferable for the structurally sampled
queries considered here. The complete results are presented in
Appendix~\ref{app:results-rw}.


\subsection{Analysis of the components of GSQ}
\label{sec:component-analysis}

Here, we study how the components of our GSQ framework contribute to
its performance. To avoid reusing test results, the analysis is based
exclusively on mean validation AE across the five folds. For each
component pair, we compare configurations that are otherwise
matched. More precisely, encoder comparisons hold the pooler and
prevalence predictor fixed; pooler comparisons hold the encoder and
predictor fixed; and predictor comparisons hold the encoder and pooler
fixed. We repeat these comparisons within two homophily groups, with
Cora-2 and Cora forming the higher-homophily group and Toloker and
Amazon the lower-homophily group, and report how frequently each
component achieves the lower validation AE. For GESN, Mean operates on
the original reservoir embeddings, whereas DeepSets and GMNet operate
on PCA-projected embeddings, following the configurations evaluated in
the experiments. Consequently, comparisons involving GESN and a
learned pooler assume the PCA transformation.


\subsubsection{APP results}

\begin{figure}[t]
  \centering
  \includegraphics[width=\linewidth]{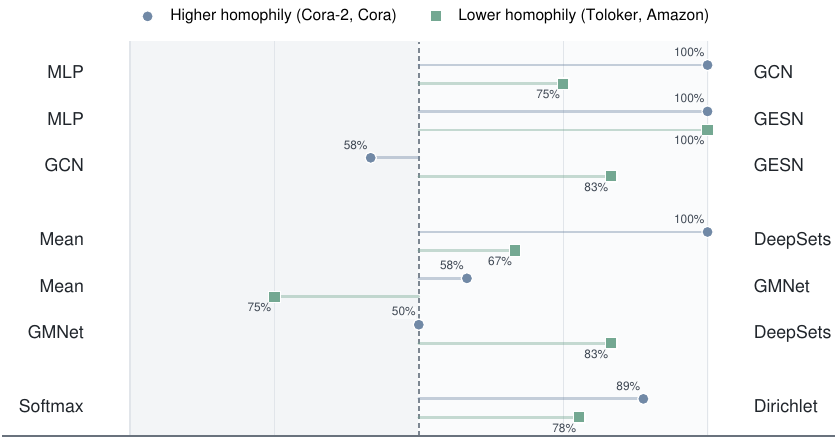}
  \caption{Pairwise comparison of the GSQ framework components under
  PPS sampling. Marker position indicates which method achieves lower
  validation AE in a larger proportion of matched configurations,
  while labels report the corresponding win rate. The dashed central
  line denotes parity. Results are reported separately for the
  higher-homophily Cora-2 and Cora graphs (blue marker) and the
  lower-homophily Toloker and Amazon graphs (green marker).}
  \label{fig:components-pps}
\end{figure}

Figure~\ref{fig:components-pps} reports the PPS comparisons. On Cora-2
and Cora, both graph-aware encoders consistently improve over MLP,
confirming that relational information is useful in this setting. The
direct comparison between GCN and GESN is less conclusive, and the
median difference in validation AE is below $.001$. The
higher-homophily results therefore do not identify a uniformly
preferable graph encoder.

The pattern is clearer on Toloker and Amazon. GESN consistently
improves over MLP and is generally preferable to GCN, with a median
reduction in validation AE of $.007$. GCN also tends to improve over
MLP, showing that relational information remains useful, but the
reservoir representation is better suited to the lower-homophily
graphs considered here.

Among the poolers, DeepSets is the most reliable choice across both
homophily groups. Its advantage is particularly clear over Mean,
whereas the comparison with GMNet is less decisive on Cora-2 and
Cora. On Toloker and Amazon, DeepSets is generally preferable to both
alternatives, while GMNet provides no consistent improvement over
Mean.

The Dirichlet predictor is also better across both graph groups. Its
median advantage over Softmax is $.010$ on Cora-2 and Cora and $.002$
on Toloker and Amazon. Overall, prior probability shift favours a
Dirichlet predictor and a learned pooler, whereas the encoder choice
is primarily determined by graph homophily. GCN and GESN are nearly
interchangeable on Cora-2 and Cora, while GESN is preferable on
Toloker and Amazon.




\subsubsection{BFP results}

\begin{figure}[t]
  \centering
  \includegraphics[width=\linewidth]{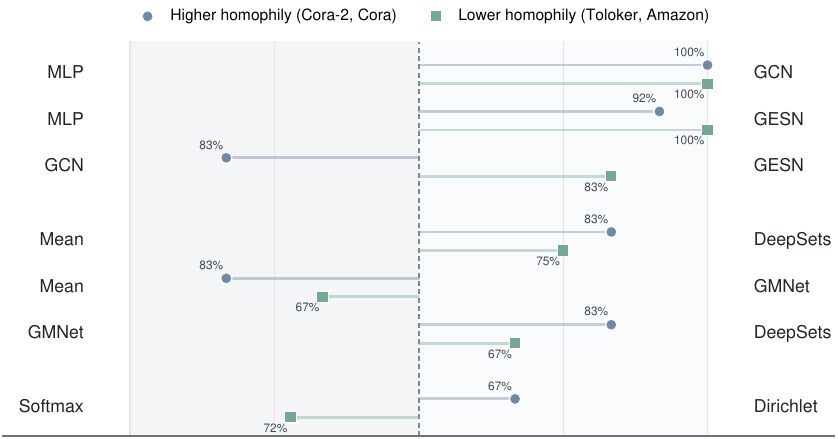}
  \caption{Pairwise comparison of the GSQ framework components under
  BFP sampling. Marker position indicates which method achieves lower
  validation AE in a larger proportion of matched configurations,
  while labels report the corresponding win rate. The dashed central
  line denotes parity. Results are reported separately for the
  higher-homophily Cora-2 and Cora graphs (blue marker) and the
  lower-homophily Toloker and Amazon graphs (green marker).}
  \label{fig:components-bfs}
\end{figure}

The breadth-first sampling results, shown in
Figure~\ref{fig:components-bfs}, exhibit a stronger interaction
between the encoder and graph homophily. GCN is the preferred encoder
on Cora-2 and Cora, whereas GESN is preferred on Toloker and
Amazon. Both graph-aware encoders improve consistently over MLP, but
the direction of their direct comparison reverses between the two
homophily groups.

DeepSets is again the most reliable pooler. It is generally preferred
to Mean and GMNet in both graph groups, although its median advantage
over Mean is only $.0003$ on Toloker and Amazon. Thus, the direction
of the comparison is consistent, but Mean remains practically
competitive on the lower-homophily graphs. GMNet provides no
systematic advantage over the other poolers.

The predictor choice is less stable. Dirichlet is generally preferred
on Cora-2 and Cora, while Softmax is more often preferred on Toloker
and Amazon. The paired differences are smaller than those observed
under prior probability shift, indicating that the predictor has less
influence than the encoder in this setting. Taken together, the
results favour GCN on the higher-homophily graphs and GESN on the
lower-homophily graphs. DeepSets provides the safest default pooler,
whereas the predictor is better selected through validation.



\subsubsection{RWP results}

\begin{figure}[t]
  \centering
  \includegraphics[width=\linewidth]{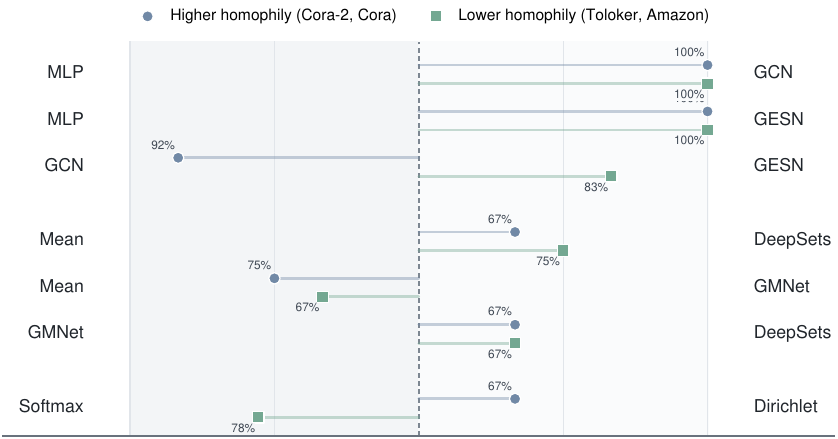}
  \caption{Pairwise comparison of the GSQ framework components under
  RWP sampling. Marker position indicates which method achieves lower
  validation AE in a larger proportion of matched configurations,
  while labels report the corresponding win rate. The dashed central
  line denotes parity. Results are reported separately for the
  higher-homophily Cora-2 and Cora graphs (blue marker) and the
  lower-homophily Toloker and Amazon graphs (green marker).}
  \label{fig:components-rw}
\end{figure}

Figure~\ref{fig:components-rw} shows that random-walk sampling
produces a pattern similar to breadth-first sampling. Both GCN and
GESN consistently outperform MLP, irrespective of homophily. Their
direct comparison is again determined by the graph group: GCN is
preferable on Cora-2 and Cora, whereas GESN is preferable on Toloker
and Amazon. The corresponding median differences are $.006$ in favour
of GCN and $.001$ in favour of GESN, respectively. The encoder
preference is therefore clear in direction, although the advantage of
GESN on the lower-homophily graphs is small.

DeepSets is generally preferable to Mean and GMNet in both groups, but
the paired differences are modest, particularly on Toloker and
Amazon. GMNet again provides no consistent advantage over Mean. The
predictor comparison follows the breadth-first pattern: Dirichlet is
more often preferred on Cora-2 and Cora, whereas Softmax is generally
preferred on Toloker and Amazon. Median predictor differences are
close to $.001$ in both groups, so this should be interpreted as a
selection tendency rather than a substantial performance difference.

Across the three protocols, the encoder produces the clearest and most
consistent effects. MLP is rarely competitive, confirming the
importance of graph-aware node representations for symmetric
quantification. GCN is generally preferable on Cora-2 and Cora,
whereas GESN is favoured on Toloker and Amazon, particularly under
structural sampling. DeepSets is the most reliable pooler, although
its advantage over Mean is often small under breadth-first and
random-walk sampling. The predictor is more protocol dependent:
Dirichlet is consistently preferred under prior probability shift,
while Softmax becomes competitive, and often preferable, on
lower-homophily graphs under structural sampling. These regularities
provide useful initial choices, but the small differences observed in
several comparisons support retaining validation-based model selection
when the graph does not fall clearly into either regime.



\subsubsection{Impact of post-hoc prevalence scaling}
\label{sec:scaling-results}

We finally assess the contribution of post-hoc prevalence scaling. For
each dataset and sampling protocol, we consider the symmetric
architecture selected by mean validation AE and inspect the scaling
alternative selected in each fold. The comparison is matched: no
scaling, scalar scaling, and vector scaling are applied to the same
base predictions and evaluated on the same fixed validation
samples. Differences between the alternatives can therefore be
attributed to the scaling transformation rather than to the underlying
model or sampled queries.

\begin{table}[!htbp]
  \centering \setlength{\tabcolsep}{7pt}
  \renewcommand{\arraystretch}{1.08}

  \caption{Post-hoc prevalence scaling selected for the final
  symmetric models. Counts indicate the number of dataset--fold
  combinations in which each alternative achieves the lowest
  validation AE. The final column reports the mean paired difference
  between the selected and unscaled validation AE, with negative
  values favouring scaling.}
  \label{tab:scaling-analysis}

\begin{tabular}{@{}lrrrr@{}}
  \toprule
  \textbf{Protocol}
  & \textbf{None}
  & \textbf{Scalar}
  & \textbf{Vector}
  & $\boldsymbol{\Delta}$\textbf{AE} \\
  \midrule
  APP & 8/20 & 6/20 & 6/20 & $-.0020$ \\
  BFP & 13/20 & 5/20 & 2/20 & $-.0007$ \\
  RWP & 15/20 & 2/20 & 3/20 & $-.0001$ \\
  \bottomrule
\end{tabular}
\end{table}

As shown in Table~\ref{tab:scaling-analysis}, scaling is most useful
under PPS. Either scalar or vector scaling is selected in 12 of the 20
dataset-fold combinations, with an average reduction in validation AE
of $0.0020$ relative to the corresponding unscaled predictions. Its
contribution is more limited under structural sampling. No scaling is
selected in 13 of 20 combinations under BFP and 15 of 20 under RWP,
while the corresponding average reductions in validation AE are
$0.0007$ and $0.0001$.

The preferred transformation also depends on the dataset. Scalar
scaling is selected in four of the five PPS folds on Toloker, whereas
vector scaling is selected in three PPS folds on Amazon. Conversely,
all Amazon models under BFP and all Toloker models under RWP retain
their original estimates. Neither transformation is therefore
uniformly preferable. These results support treating scaling as an
optional validation-selected step and retaining the identity
transformation as a candidate, particularly under structural sampling,
where the unscaled estimates are generally preferred. A per-dataset
breakdown of these results is presented in
Appendix~\ref{app:scaling-results}.




\section{Related work}
\label{sec:related-work}
%

\subsection{Quantification for linked data} While the literature on
quantification is now substantial (see \citep{Esuli:2023os} for a
fairly recent overview), the papers that have addressed quantification
for linked data are just a handful.

The earliest such work is~\citep{Tang:2010uq}, where the authors
propose two methods. The first (dubbed CBQ, for ``classification-based
quantification'') consists of individually classifying each unlabelled
node of the graph via a relational classifier (i.e., a classifier
capable of exploiting both the content of a node and its surrounding
link structure) and then correcting the resulting prevalence estimates
via the ACC (non-relational) quantification method. The second method
(dubbed LBQ, for ``link-based quantification'') consists of estimating
the fraction of nodes $k$ hops away from an anchor node $v$ as a
mixture of class-conditional such fractions; since all these fractions
are either available or can be estimated from the training data, the
parameters of the mixture (i.e., the class prevalences) can be
estimated via standard algebra, after which the final prevalence
estimate is computed as the median across the estimates obtained for
each anchor node $v$. However, while CBQ can be experimentally
compared with the method proposed in the present paper, LBQ cannot,
since it allows estimating class prevalence in the entire graph only,
and not in random samples, as our method (and all the other methods
discussed in this section) instead allows.

This work was followed by~\citep{Milli:2015mz}, which proposes two
aggregative methods, ``Community Discovery for Quantification'' (CDQ)
and ``Ego Networks for Quantification'' (ENQ). Both methods first
assign classes to unlabelled nodes using the network structure (CDQ
via community detection, and ENQ via ego networks), and then correct
the resulting estimates. To this end, CDQ uses frequency or density
within communities, while ENQ uses the majority class in a node's ego
network, with isolated nodes defaulting to the training
distribution. Both methods ignore node features and assume homophily.

The other methods we survey are radically more recent.
\citet{Micheli:2025ez} introduce XNQ, a method that combines
unsupervised node embeddings computed by randomised recursive graph
neural networks with \citet{Saerens:2002uq}'s (non-relational)
quantification method, and show that, under prior probability shift,
XNQ largely outperforms~\citet{Tang:2010uq}'s CBQ
and~\citet{Milli:2015mz}'s CDQ and ENQ. \citet{Damke:2025cr,
Damke:2025qf} propose a quantification method for graph nodes that
explicitly addresses a type of shift different from prior probability
shift, which the authors call \textit{structural covariate shift},
defined as the situation in which the training data come from a
different region of the graph than the test data. The method that they
propose is based on ``structural importance sampling''; while
\citep{Damke:2025cr} adapts to structural covariate shift the ACC
quantification method~\citep{Forman:2008kx}, \citep{Damke:2025qf} does
the same with the (better performing) KDEy
method~\citep{Moreo:2025lq}.
Note that \citep{Milli:2015mz, Tang:2010uq}, similarly to other early
quantification works, do not mention dataset shift, and do not state
which type of dataset shift they address. The analysis of dataset
shift is instead central to~\citep{Damke:2025cr, Damke:2025qf,
Micheli:2025ez}.

\subsection{Non-aggregative quantification and symmetric
quantification} 
In contrast to aggregative quantifiers, non-aggregative quantifiers do
not rely on a classifier issuing label predictions for the test
instances as an intermediate step towards class prevalence
estimation. Representative examples of this family include
ReadMe~\citep{Hopkins:2010fk}, ReadMe2~\citep{Jerzak:2022pt},
HDx~\citep{Gonzalez-Castro:2013fk}, and
EDx~\citep{Kawakubo:2016nm}. Beyond being non-aggregative, these
methods share a common trait, namely, they belong to the family of
distribution-matching methods. That is, they estimate class
prevalences by solving an optimisation problem in which the prevalence
values are adjusted so that a mixture of class-conditional
distributions best matches the distribution observed in the test
data~\citep{Firat:2016uq}. Different methods adopt different
representations of the underlying distributions and different matching
criteria. For instance, ReadMe and EDx operate directly on the
original covariates, ReadMe2 relies on document embeddings, and HDx
relies on per-feature histograms. Likewise, ReadMe and ReadMe2
formulate the matching problem as a least-squares regression task, HDx
relies on the Hellinger Distance, and EDx minimises the expected
Energy Distance between distributions. Note that, although a
classifier is not trained as part of the quantification process, all
methods described above still require instance-level labels in the
training set.

Non-aggregativity is appealing because, in the spirit of Vapnik's
principle (see the Introduction), the available information is used
directly to solve the target problem, without resorting to a more
general task (namely, classification). Despite this appealing
property, recent empirical evaluations have consistently shown that
non-aggregative quantifiers tend to underperform aggregative ones.

However, in order to tackle settings in which supervision is available
only at the sample level, i.e., where training samples are labelled
exclusively by prevalence and individual labels are never observed,
one requires not only non-aggregativity, but symmetry, in the sense
discussed in \S\ref{sec:scenariosandmethods}.  \textit{Symmetric}
quantification methods deviate from the traditional asymmetric
character of quantification (training items are individuals,
unlabelled items are samples of individuals) in that both their
training items and the items they need to label are samples. These
methods are useful, especially since in scenarios where privacy
constraints dictate that no individual-level inference should be made,
individual-level labels for training are unlikely to be available
either. It is precisely in this direction that recent neural
approaches have emerged.

In recent years, several neural architectures specifically designed to
process set-valued data have been proposed. Although these
architectures were not originally devised for quantification, they are
naturally suited to prevalence estimation because they are built upon
permutation-invariant (unbiased towards ordering) and variable-size
(unbiased towards sample size) representations.\footnote{A function
$f$ is said to be permutation-invariant if $f(\sigma)=f(\pi(\sigma))$
for any permutation $\pi$ of the input sequence $\sigma$.}

The first of these architectures is DeepSets~\citep{zaheer2017deep},
that relies on permutation-invariant pooling operators such as max,
average, or median, to aggregate instances into a fixed-size
representation. \citet{Qi:2020qp} proposed DQN, which adapts DeepSets
to quantification problems. A more expressive alternative was proposed
by \citet{lee2019set}, which replaced the simple pooling operators of
DeepSets with transformer-based attention mechanisms capable of
modelling interactions among the elements of the set. The method,
called ``Set Transformers'', uses the attention mechanism of
transformers omitting positional encodings (since the order of the
instances is irrelevant). In order to reduce the combinatorial
complexity, attention is computed with respect to a reduced set of
learnable \emph{inducing points} which act as latent representatives
of the input space.

More recently, \citet{Perez-Mon:2025jt} proposed HistNetQ, which can
be viewed as a neural counterpart of HDx. HistNetQ replaces
traditional histograms with learnable histogram layers, where both the
bin centres and bin widths are trainable parameters and the histogram
operator itself is differentiable. Unlike classical non-aggregative
quantifiers, HistNetQ was specifically designed for scenarios in which
training samples are labelled only by prevalence and instance-level
labels are unavailable. To mitigate the severe data scarcity that
typically arises in this setting, HistNetQ introduces a sample-mixing
strategy that generates additional training samples by combining
existing ones and inferring silver prevalence labels from the
resulting mixtures. As in HDx, however, HistNetQ ultimately relies on
a factorised representation of the feature space through per-dimension
histograms, which inevitably assumes independence among feature
dimensions. To address this limitation, \citet{Perez-Mon:2026rd}
proposed GMNet, which replaces histogram-based pooling with learnable
multivariate Gaussian mixtures capable of modelling dependencies
between feature dimensions.


%
%


\section{Conclusions}
\label{sec:conclusion}

This article has introduced GSQ, a framework for graph node
quantification.
GSQ enforces \textit{symmetric} quantification, i.e., the labelled
items used for training and validation, and the unlabelled items that
are the object of inference, are of the same type (i.e., sets of
nodes), with no classifier for labelling individual datapoints being
involved in the pipeline; this is advantageous, since GSQ can also
address application scenarios in which individual labelled datapoints
are not available and/or inferences at the individual datapoint level
are undesirable. The framework combines interchangeable node encoders,
permutation-invariant poolers, prevalence predictors, and optional
post-hoc scaling, allowing to seamlessly tackle binary and multi-class
quantification tasks.

Our experiments show that graph node symmetric quantification
constitutes a strong practical alternative to established aggregative
node quantification methods. Under prior probability shift, GSQ
remains competitive with XNQ, a strong baseline specifically designed
for this setting, and outperforms it on the Amazon dataset. Under
structural covariate shift, the advantage of direct sample-level
learning becomes clearer: GSQ achieves a lower mean test error in
seven of the eight comparisons generated by the BFP and RWP protocols,
and is substantially tied in the remaining one. The component analysis
further demonstrates that the proposed framework is effective across
different graphs and shift conditions, while also providing
practitioners with guidance on selecting its components according to
the characteristics of the input graph and the type of shift expected
to affect it.

This study is not without limitations. The main limitation (which is
not specific to our contribution, but applies more generally to
symmetric quantification methods) is that the evaluation we have
carried out relies on controlled protocols to generate node samples
(both for training and for testing purposes) and the associated
dataset shifts. In particular, although GSQ requires only sample-level
supervision, the labelled node samples we use in our experiments are
generated, via shift generation protocols, from individually labelled
nodes; it is unclear whether the accuracy results obtained in our
experiments could be obtained also in scenarios (such as the one of
Example~\ref{ex:example1}) where supervision is \textit{naturally}
available only at the node sample level, and in which labelled samples
cannot thus be generated at will. While shift generation protocols
allow symmetric quantification methods to be extensively benchmarked,
further insights can be gained from datasets that natively provide
labelled aggregate samples and naturally occurring shifts in
real-world applications.

These limitations suggest several promising directions for future
work. Connected to the latter limitation described above, a
particularly important step will be to evaluate GSQ in applications
that provide genuinely aggregate supervision and naturally occurring
graph queries, thereby testing symmetric quantification without
relying on individual labels even to construct training
samples. Extending the framework to inductive (rather than
transductive) and evolving-graph settings would allow it to process
nodes and graphs that were not observed during training. The
interactions observed among graph structure, sampling protocol, and
architectural choices also motivate adaptive models that select or
combine encoders and poolers using observable graph and shift
characteristics.  In a further extension of this work's scope, one
could investigate quantification on dynamic graphs, where structural
intra-graph (at node or sub-graph level) or inter-graph dataset shifts
are caused by the temporal evolution of node relationships, e.g., in
epidemic diffusion processes \citep{Micheli2022}.  Finally, a
theoretical analysis of identifiability, sample complexity, and
robustness under different graph-sampling mechanisms would complement
the empirical evidence presented here.

Overall, our results establish graph node symmetric quantification as
a viable, flexible, and proficient alternative to the aggregative
paradigm, and especially so under structurally induced dataset shift.

\backmatter



\bmhead{Acknowledgements} A.\ Micheli, M.\ Podda, and D.\ Tortorella
were partly supported by Project DEEP-GRAPH, funded by the Italian
Ministry of University and Research, PRIN 2022 (project code:
2022YLRBTT, CUP: I53C24002440006), and by Project PAN-HUB, funded by
the Italian Ministry of Health (POS 2014--2020, project ID: T4-AN-07,
CUP:
I53C22001300001). 

\bmhead{Declarations}

The authors declare the following:

\begin{itemize}
\item \textsf{Conflict of interest/Competing interests.} The authors
  have no relevant financial or non-financial interests to disclose.
\item \textsf{Code availability and data availability:} All the code
  and data that are necessary to replicate the results reported in
  this paper are available from
  \url{https://github.com/marcopodda/gsq}
\end{itemize}




\newpage

\begin{appendices}


\section{Additional experimental details}\label{app:additional-details-experiments}
This appendix provides further details on the experimental setup used
to assess the aggregative baselines and the proposed framework,
respectively.


\subsection{Hyperparameters of the aggregative baselines}
Table~\ref{tab:aggregative-hyperparameters} details the different
hyperparameters used to train the aggregative baselines.

\begin{table}[!htbp]
  \centering \footnotesize \setlength{\tabcolsep}{5pt}
  \renewcommand{\arraystretch}{1.12}
  \caption{Aggregative model configurations. Sets denote search grids;
  other settings are fixed. Here, $d$ is the number of node features,
  $d_h$ is the embedding dimension, and $C=|\mathcal{Y}|$. Shared
  training settings follow Table~\ref{tab:framework-hyperparameters}.}
  \label{tab:aggregative-hyperparameters}

 \begin{tabular}{@{}lll@{}}
   \toprule
   & \textbf{Parameter} & \textbf{Values} \\
   \midrule

   \rowcolor{black!8}
   \multicolumn{3}{@{}l@{}}{\strut\textbf{XNQ}} \\
   \addlinespace[3pt]

   \multirow[c]{6}{*}{GESN}
   & Reservoir dimension $d_h$
                        & $\{1024,2048\}$ \\
   & Input scale $\alpha_{\mathrm{in}}$
                        & $\{0.1,0.5,1\}$ \\
   & Recurrent scale $\alpha_{\mathrm{rec}}$
                        & $\{1,10,100\}$ \\
   & Iteration setting
                        & $30$ \\
   \addlinespace[5pt]

   \multirow[c]{2}{*}{Node classifier}
   & Mapping
                        & Linear $d_h\to C$, followed by softmax \\
   & Objective
                        & Node-level cross-entropy \\
   \addlinespace[5pt]

   \multirow[c]{4}{*}{SLD}
   & Initial prevalence
                        & Training-partition class proportions \\
   & Iteration limit
                        & $1{,}000$ \\
   & Convergence tolerance
                        & Mean absolute prevalence change below $10^{-4}$ \\
   & Numerical floor
                        & $10^{-4}$ \\
   \addlinespace[5pt]

   \multirow[c]{3}{*}{Scaling}
   & Alternatives
                        & None, scalar, vector \\
   & Fitting
                        & Node probabilities and scaling-node labels \\
   & Selection criterion
                        & Downstream validation AE \\
   \midrule

   \rowcolor{black!8}
   \multicolumn{3}{@{}l@{}}{\strut\textbf{APPNP-KDEy}} \\
   \addlinespace[3pt]

   \multirow[c]{6}{*}{APPNP}
   & Feature network
                        & $d\to64\to64$ \\
   & Activation; dropout
                        & ReLU; $0.1$ after each linear layer \\
   & Propagation steps
                        & $10$ \\
   & Restart probability
                        & $0.1$ \\
   & Propagation dropout
                        & $0$ \\
   & Self-loops
                        & Enabled \\
   \addlinespace[5pt]

   \multirow[c]{3}{*}{Node classifier}
   & Mapping
                        & Linear $64\to C$, followed by softmax \\
   & Objective
                        & Node-level cross-entropy \\
   \addlinespace[5pt]

   \multirow[c]{3}{*}{KDEy}
   & Adjustment statistics
                        & Class-conditional mean probabilities \\
   & Adjustment solver
                        & Simplex-constrained minimisation \\
   \midrule

   \rowcolor{black!8}
   \multicolumn{3}{@{}l@{}}{\strut\textbf{Structural correction (SIS)}} \\
   \addlinespace[3pt]

   \multirow[c]{5}{*}{KDEy correction}
   & Structural kernel
                        & Finite personalised PageRank diffusion \\
   & Diffusion steps
                        & $10$ \\
   & Restart probability
                        & $0.1$ \\
   & Interpolation weight
                        & $\{0,0.5,0.9,1\}$ for BFP/RWP; $0$ for PPS \\
   & Selection criterion
                        & Validation AE \\
   \midrule

   \rowcolor{black!8}
   \multicolumn{3}{@{}l@{}}{\strut\textbf{Shared settings}} \\
   \addlinespace[3pt]

   Training
   & Optimisation and stopping
                        & As in Table~\ref{tab:framework-hyperparameters} \\
   \bottomrule
 \end{tabular}
\end{table}
\FloatBarrier


\subsection{Hyperparameters of the symmetric framework}
Table~\ref{tab:framework-hyperparameters} details the different
configurations of the proposed symmetric framework assessed in the
experiments.

\begin{table}[!htbp]
  \centering \footnotesize \setlength{\tabcolsep}{5pt}
  \renewcommand{\arraystretch}{1.12}
  \caption{Non-aggregative framework configurations. Sets denote
  search grids; other settings are fixed. Here, $d$ is the number of
  node features and $C=|\mathcal{Y}|$. The embedding dimension $d_h$
  is kept fixed during layer processing.}
  \label{tab:framework-hyperparameters}

 \begin{tabular}{
 @{}>{\raggedright\arraybackslash}p{0.19\textwidth}
 >{\raggedright\arraybackslash}p{0.30\textwidth}
 >{\raggedright\arraybackslash}p{\dimexpr0.51\textwidth-20pt\relax}@{}
 }
   \toprule
   & \textbf{Parameter} & \textbf{Values} \\
   \midrule

   \rowcolor{black!8}
   \multicolumn{3}{@{}l@{}}{\strut\textbf{Node encoders}} \\
   \addlinespace[3pt]

   \multirow[c]{3}{=}{MLP / GCN}
   & Hidden dimension $d_h$; depth $L$
                        & $\{64,128\}$; $\{2,3\}$ \\
   & Input projection
                        & Linear $d\to d_h$, followed by ReLU \\
   & Hidden-layer processing
                        & LayerNorm, ReLU, dropout $0.1$ \\
   \addlinespace[5pt]

   \multirow[c]{7}{=}{GESN}
   & Iterations
                        & $30$ \\
   & Reservoir dimension
                        & $\{1024,2048\}$ \\
   & Input scale $\alpha_{\mathrm{in}}$
                        & $\{0.1,0.5,1\}$ \\
   & Recurrent scale $\alpha_{\mathrm{rec}}$
                        & $\{1,10,100\}$ \\
   & PCA
                        & $(d_{\mathrm{GESN}},d_{\mathrm{PCA}})
                          \in\{(1024,64),(2048,128)\}$ \\
   \midrule

   \rowcolor{black!8}
   \multicolumn{3}{@{}l@{}}{\strut\textbf{Poolers}} \\
   \addlinespace[3pt]

   Mean
   & --
                        & -- \\
   \addlinespace[5pt]

   DeepSets
   & Node and sample MLPs
                        & $d_h\to d_h\to d_h$, ReLU between linear layers \\
   \addlinespace[5pt]

   \multirow[c]{6}{=}{GMNet}
   & Views; Gaussians per view
                        & $9$; $100$ \\
   & View dimensions
                        & $d_h\to50\to5$ \\
   & View activations; dropout
                        & LeakyReLU (slope $0.01$), sigmoid; $0.5$ \\
   & Covariance
                        & Full; Cholesky diagonal uses softplus \\
   & Initial Cholesky factor
                        & $0.1\mathbf{I}$ \\
   & Reduction; output projection
                        & Mean density responses; linear $900\to d_h$ \\
   \midrule

   \rowcolor{black!8}
   \multicolumn{3}{@{}l@{}}{\strut\textbf{Prevalence Predictors}} \\
   \addlinespace[3pt]

   \multirow[c]{2}{=}{Softmax head}
   & Mapping
                        & Linear $d_h\to C$, followed by softmax \\
   & Objective
                        & MSE averaged over samples and classes \\
   \addlinespace[5pt]

   \multirow[c]{3}{=}{Dirichlet head}
   & Mapping
                        & Linear $d_h\to C$, followed by softplus \\
   & Target smoothing
                        & Clamp at $10^{-4}$ and renormalise \\
   & Concentration penalty
                        & $\lambda_{\alpha}(\log\alpha_{\sigma,0})^2$,
                          with $\lambda_{\alpha}=10^{-3}$ \\
   \midrule

   \rowcolor{black!8}
   \multicolumn{3}{@{}l@{}}{\strut\textbf{Post-hoc scaling}} \\
   \addlinespace[3pt]

   \multirow[c]{6}{=}{Scaling}
   & Alternatives
                        & None, scalar, vector \\
   & Initialisation
                        & Unit temperatures and zero biases \\
   & Objective; regularisation
                        & Cross-entropy; weight $10^{-3}$ \\
   & Optimiser
                        & L-BFGS, learning rate $0.1$, Wolfe line search \\
   & Iteration limit
                        & $100$ \\
   & Selection criterion
                        & Validation AE \\
   \midrule

   \rowcolor{black!8}
   \multicolumn{3}{@{}l@{}}{\strut\textbf{Training}} \\
   \addlinespace[3pt]

   \multirow[c]{6}{=}{Shared settings}
   & Optimiser
                        & AdamW, learning rate $10^{-3}$,
                          weight decay $10^{-5}$ \\
   & Gradient clipping
                        & Value clipping at $1$ \\
   & Epochs; validation frequency
                        & Minimum $20$, maximum $500$; every $10$ epochs \\
   & Learning-rate scheduler
                        & Factor $0.5$, patience $5$,
                          min. learning rate $10^{-6}$ \\
   & Early stopping
                        & Patience $10$ validation checks \\
   & Improvement threshold
                        & Absolute validation AE improvement
                          of $5\times10^{-5}$ \\
   \bottomrule
 \end{tabular}
\end{table}


\section{Additional results}\label{app:additional-results}
Here, we report for completeness the detailed results obtained in our
experiments.


\subsection{Complete results using the APP sampling protocol}\label{app:results-pps}
Table~\ref{tab:results-pps} reports the complete set of results
obtained for the different variants examined in this study for the APP
sampling protocol. For each encoder/pooler/head combination, we select
the best hyperparameter configuration and assess their performance on
the test set through the mean AE metric.

\begin{table}[!htbp]
  \centering \footnotesize \setlength{\tabcolsep}{4pt}
  \renewcommand{\arraystretch}{1.08}

  \caption{Test absolute error under the APP sampling protocol.
  Entries report mean AE (standard deviation) across the five folds.
  Entries in \textbf{boldface} identify the aggregative and symmetric
  configurations selected independently by lowest mean validation AE.}
  \label{tab:results-pps}

\begin{tabularx}{\textwidth}{
 @{}llc *{4}{>{\centering\arraybackslash}X} @{}
}
\toprule \textbf{Encoder} & \textbf{Pooler / method} & \textbf{Head}
& \textbf{Cora-2} & \textbf{Toloker} & \textbf{Cora} & \textbf{Amazon} \\
\midrule
\multirow{2}{*}{Aggregative} & XNQ & -- & \result{\textbf{.011}}{.005} & \result{\textbf{.028}}{.004} & \result{\textbf{.011}}{.002} & \result{\textbf{.101}}{.046} \\[3pt]
& APPNP-KDEy & -- & \result{.012}{.009} & \result{.046}{.006} &
\result{.013}{.003}
& \result{.153}{.028}\\
\midrule
\multirow{6}{*}{MLP} & \multirow{2}{*}{Mean} & Softmax & \result{.039}{.007} & \result{.050}{.004} & \result{.035}{.001} & \result{.083}{.006} \\
& & Dirichlet & \result{.041}{.011} & \result{.054}{.008} & \result{.028}{.003} & \result{.077}{.012} \\[3pt]
& \multirow{2}{*}{DeepSets} & Softmax & \result{.040}{.016} & \result{.042}{.007} & \result{.026}{.002} & \result{.116}{.006} \\
& & Dirichlet & \result{.038}{.011} & \result{.044}{.009} & \result{.026}{.002} & \result{.109}{.006} \\[3pt]
& \multirow{2}{*}{GMNet} & Softmax & \result{.044}{.006} & \result{.048}{.005} & \result{.040}{.004} & \result{.131}{.000} \\
& & Dirichlet & \result{.047}{.028} & \result{.049}{.009} & \result{.027}{.004} & \result{.131}{.000} \\
\midrule
\multirow{6}{*}{GCN} & \multirow{2}{*}{Mean} & Softmax & \result{.029}{.003} & \result{.045}{.005} & \result{.030}{.002} & \result{.081}{.004} \\
& & Dirichlet & \result{.021}{.008} & \result{.040}{.004} & \result{.020}{.003} & \result{.078}{.005} \\[3pt]
& \multirow{2}{*}{DeepSets} & Softmax & \result{.013}{.006} & \result{.038}{.003} & \result{.018}{.003} & \result{.094}{.007} \\
& & Dirichlet & \result{.014}{.009} & \result{.036}{.003} & \result{.017}{.004} & \result{.097}{.008} \\[3pt]
& \multirow{2}{*}{GMNet} & Softmax & \result{.034}{.005} & \result{.047}{.005} & \result{.032}{.001} & \result{.131}{.000} \\
& & Dirichlet & \result{\textbf{.012}}{.006} & \result{.038}{.002} & \result{\textbf{.014}}{.003} & \result{.131}{.000} \\
\midrule
\multirow{6}{*}{GESN} & \multirow{2}{*}{Mean} & Softmax & \result{.100}{.089} & \result{.048}{.008} & \result{.033}{.002} & \result{.128}{.032} \\
& & Dirichlet & \result{.021}{.007} & \result{.040}{.003} & \result{.019}{.001} & \result{.112}{.031} \\[3pt]
& \multirow{2}{*}{DeepSets} & Softmax & \result{.020}{.006} & \result{.032}{.007} & \result{.017}{.002} & \result{.079}{.001} \\
& & Dirichlet & \result{.017}{.008} & \result{\textbf{.030}}{.006} & \result{.018}{.002} & \result{.075}{.005} \\[3pt]
& \multirow{2}{*}{GMNet} & Softmax & \result{.043}{.013} & \result{.045}{.002} & \result{.038}{.003} & \result{.074}{.009} \\
& & Dirichlet & \result{.015}{.007} & \result{.031}{.005} & \result{.016}{.003} & \result{\textbf{.065}}{.003} \\
\bottomrule
\end{tabularx}
\end{table}


\subsubsection{Diagonal plots under PPS}
\label{app:pps-diagonal}

\begin{figure}[t]
  \centering
  \begin{subfigure}[t]{0.48\textwidth}
    \centering
    \includegraphics[width=\linewidth]{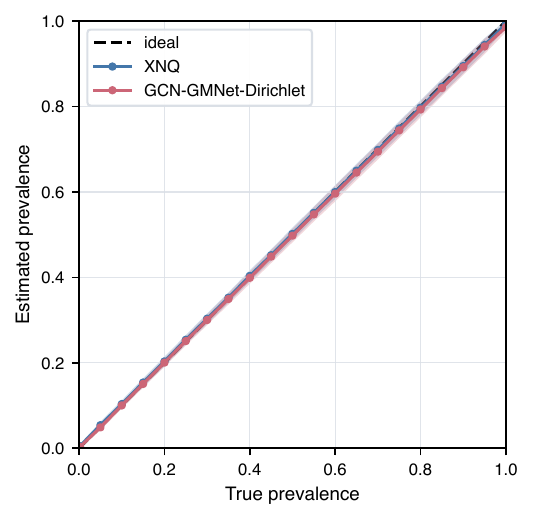}
    \caption{Cora-2}
    \label{fig:pps-diagonal-cora2}
  \end{subfigure}
  \hfill
  \begin{subfigure}[t]{0.48\textwidth}
    \centering
    \includegraphics[width=\linewidth]{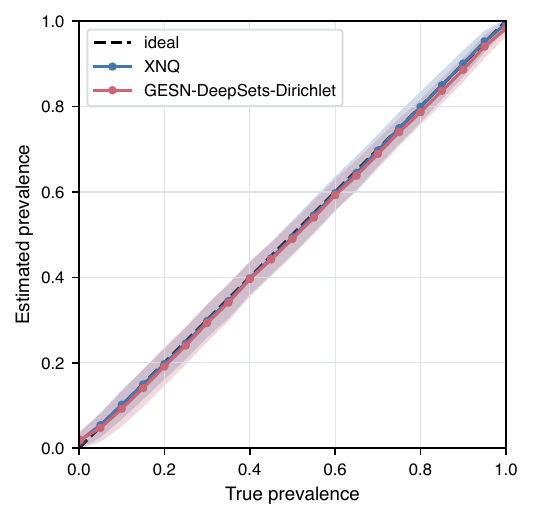}
    \caption{Toloker}
    \label{fig:pps-diagonal-toloker}
  \end{subfigure}
 \caption{Diagonal plots under PPS for the binary datasets. Each panel compares the validation-selected aggregative and symmetric configurations. Curves report the mean estimated positive-class prevalence at each true prevalence, while shaded regions denote one standard deviation. For GESN--DeepSets--Dirichlet, the reservoir embeddings are PCA-projected before pooling. 
 }
 \label{fig:pps-diagonal}
\end{figure}

Figure~\ref{fig:pps-diagonal} reports diagonal plots for the binary
quantification tasks under PPS. A diagonal plot relates the estimated
prevalence of the positive class to its true prevalence. An ideal
quantifier follows the main diagonal across the entire prevalence
range, while systematic deviations indicate prevalence-dependent
bias. We restrict this analysis to binary tasks because they can be
represented by a single class prevalence; in the multi-class case,
separate class-wise plots would ignore the dependence among the
components of the prevalence vector. We consider only APP as sampling
protocol because its controlled prevalence grid provides observations
across the full prevalence range, allowing the behaviour of the
methods to be compared at the same target values. By contrast, BFP and
RWP generate prevalences indirectly through graph traversal and
therefore provide an uneven coverage of this range.

On Cora-2, both XNQ and GCN--GMNet--Dirichlet remain close to the
ideal diagonal throughout the prevalence range, with only minor
deviations at high prevalences. On Toloker, both methods exhibit
greater variability and a mild tendency to underestimate
prevalence. XNQ remains closer to the diagonal overall, consistently
with its lower test AE under PPS. Thus, the aggregate results reported
in the main text do not conceal pronounced failures in specific
regions of the prevalence range.


\subsection{Complete results using the BFP sampling protocol}\label{app:results-bfs}
Table~\ref{tab:results-bfs} reports the complete set of results
obtained for the different variants examined in this study for the BFP
sampling protocol. For each encoder/pooler/head combination, we select
the best hyperparameter configuration and assess their performance on
the test set through the mean AE metric.

\begin{table}[!htbp]
  \centering \footnotesize \setlength{\tabcolsep}{4pt}
  \renewcommand{\arraystretch}{1.08}

  \caption{Test absolute error under the BFP sampling protocol.
  Entries report mean AE (standard deviation) across the five folds.
  Entries in \textbf{boldface} identify the aggregative and symmetric
  configurations selected independently by lowest mean validation AE.}
  \label{tab:results-bfs}

\begin{tabularx}{\textwidth}{
 @{}llc *{4}{>{\centering\arraybackslash}X} @{}
}
\toprule \textbf{Encoder} & \textbf{Pooler / method} & \textbf{Head}
& \textbf{Cora-2} & \textbf{Toloker} & \textbf{Cora} & \textbf{Amazon} \\
\midrule
\multirow{2}{*}{Aggregative} & XNQ & -- & \result{\textbf{.047}}{.032} & \result{.400}{.028} & \result{.030}{.004} & \result{\textbf{.130}}{.011} \\[3pt]
& APPNP-KDEy & -- & \result{.029}{.017} & \result{\textbf{.262}}{.017}
& \result{\textbf{.026}}{.003}
& \result{.195}{.024} \\
\midrule
\multirow{6}{*}{MLP} & \multirow{2}{*}{Mean} & Softmax & \result{.070}{.023} & \result{.045}{.009} & \result{.042}{.003} & \result{.040}{.003} \\
& & Dirichlet & \result{.088}{.035} & \result{.046}{.006} & \result{.035}{.007} & \result{.043}{.003} \\[3pt]
& \multirow{2}{*}{DeepSets} & Softmax & \result{.049}{.024} & \result{.045}{.011} & \result{.040}{.001} & \result{.041}{.003} \\
& & Dirichlet & \result{.051}{.017} & \result{.045}{.008} & \result{.048}{.005} & \result{.041}{.003} \\[3pt]
& \multirow{2}{*}{GMNet} & Softmax & \result{.086}{.038} & \result{.081}{.009} & \result{.041}{.002} & \result{.041}{.003} \\
& & Dirichlet & \result{.055}{.025} & \result{.049}{.015} & \result{.048}{.007} & \result{.042}{.005} \\
\midrule
\multirow{6}{*}{GCN} & \multirow{2}{*}{Mean} & Softmax & \result{.035}{.004} & \result{.044}{.007} & \result{.029}{.002} & \result{.036}{.002} \\
& & Dirichlet & \result{.024}{.010} & \result{.042}{.012} & \result{.026}{.001} & \result{.036}{.002} \\[3pt]
& \multirow{2}{*}{DeepSets} & Softmax & \result{.018}{.005} & \result{.046}{.009} & \result{\textbf{.025}}{.002} & \result{.036}{.001} \\
& & Dirichlet & \result{\textbf{.014}}{.006} & \result{.043}{.012} & \result{.030}{.004} & \result{.037}{.003} \\[3pt]
& \multirow{2}{*}{GMNet} & Softmax & \result{.045}{.020} & \result{.054}{.017} & \result{.034}{.002} & \result{.040}{.003} \\
& & Dirichlet & \result{.016}{.009} & \result{.044}{.012} & \result{.032}{.009} & \result{.038}{.001} \\
\midrule
\multirow{6}{*}{GESN} & \multirow{2}{*}{Mean} & Softmax & \result{.080}{.066} & \result{.037}{.009} & \result{.035}{.005} & \result{.050}{.011} \\
& & Dirichlet & \result{.083}{.081} & \result{.042}{.009} & \result{.035}{.006} & \result{.042}{.007} \\[3pt]
& \multirow{2}{*}{DeepSets} & Softmax & \result{.047}{.027} & \result{.034}{.011} & \result{.029}{.006} & \result{\textbf{.036}}{.001} \\
& & Dirichlet & \result{.042}{.036} & \result{.035}{.009} & \result{.032}{.004} & \result{.035}{.002} \\[3pt]
& \multirow{2}{*}{GMNet} & Softmax & \result{.071}{.025} & \result{.047}{.013} & \result{.047}{.006} & \result{.041}{.003} \\
& & Dirichlet & \result{.051}{.019} & \result{\textbf{.034}}{.009} & \result{.044}{.007} & \result{.042}{.003} \\
\bottomrule
\end{tabularx}
\end{table}


\subsection{Complete results using the RWP sampling protocol}\label{app:results-rw}
Table~\ref{tab:results-rw} reports the complete set of results
obtained for the different variants examined in this study for the RWP
sampling protocol. For each encoder/pooler/head combination, we select
the best hyperparameter configuration and assess their performance on
the test set through the mean AE metric.

\begin{table}[!htbp]
  \centering \footnotesize \setlength{\tabcolsep}{4pt}
  \renewcommand{\arraystretch}{1.08}

  \caption{Test absolute error under the RWP sampling protocol.
  Entries report mean AE (standard deviation) across the five folds.
  Entries in \textbf{boldface} identify the aggregative and symmetric
  configurations selected independently by lowest mean validation AE.}
  \label{tab:results-rw}

\begin{tabularx}{\textwidth}{
 @{}llc *{4}{>{\centering\arraybackslash}X} @{}
}
\toprule \textbf{Encoder} & \textbf{Pooler / method} & \textbf{Head}
& \textbf{Cora-2} & \textbf{Toloker} & \textbf{Cora} & \textbf{Amazon} \\
\midrule
\multirow{2}{*}{Aggregative} & XNQ & -- & \result{{.024}}{.024} & \result{.222}{.022} & \result{{.027}}{.006} & \result{\textbf{.156}}{.010} \\[3pt]
& APPNP-KDEy & -- & \result{\textbf{.014}}{.002} &
\result{\textbf{.212}}{.022} & \result{\textbf{.022}}{.002}
& \result{.202}{.024} \\
\midrule
\multirow{6}{*}{MLP} & \multirow{2}{*}{Mean} & Softmax & \result{.040}{.010} & \result{.036}{.001} & \result{.034}{.003} & \result{.051}{.004} \\
& & Dirichlet & \result{.028}{.012} & \result{.039}{.003} & \result{.031}{.002} & \result{.054}{.004} \\[3pt]
& \multirow{2}{*}{DeepSets} & Softmax & \result{.031}{.010} & \result{.036}{.002} & \result{.033}{.004} & \result{.050}{.004} \\
& & Dirichlet & \result{.025}{.005} & \result{.038}{.003} & \result{.043}{.006} & \result{.054}{.005} \\[3pt]
& \multirow{2}{*}{GMNet} & Softmax & \result{.144}{.015} & \result{.068}{.010} & \result{.041}{.003} & \result{.052}{.003} \\
& & Dirichlet & \result{.029}{.014} & \result{.037}{.002} & \result{.040}{.006} & \result{.054}{.003} \\
\midrule
\multirow{6}{*}{GCN} & \multirow{2}{*}{Mean} & Softmax & \result{.025}{.004} & \result{.036}{.001} & \result{.023}{.001} & \result{.045}{.003} \\
& & Dirichlet & \result{.014}{.004} & \result{.035}{.003} & \result{.022}{.002} & \result{.047}{.003} \\[3pt]
& \multirow{2}{*}{DeepSets} & Softmax & \result{.012}{.004} & \result{.035}{.002} & \result{\textbf{.022}}{.002} & \result{.046}{.003} \\
& & Dirichlet & \result{.011}{.003} & \result{.036}{.001} & \result{.025}{.003} & \result{.049}{.004} \\[3pt]
& \multirow{2}{*}{GMNet} & Softmax & \result{.132}{.056} & \result{.042}{.003} & \result{.027}{.002} & \result{.047}{.002} \\
& & Dirichlet & \result{\textbf{.011}}{.004} & \result{.035}{.003} & \result{.026}{.004} & \result{.049}{.003} \\
\midrule
\multirow{6}{*}{GESN} & \multirow{2}{*}{Mean} & Softmax & \result{.042}{.018} & \result{.036}{.003} & \result{.033}{.002} & \result{.063}{.017} \\
& & Dirichlet & \result{.040}{.024} & \result{.036}{.002} & \result{.032}{.005} & \result{.063}{.009} \\[3pt]
& \multirow{2}{*}{DeepSets} & Softmax & \result{.017}{.009} & \result{\textbf{.034}}{.003} & \result{.025}{.001} & \result{\textbf{.046}}{.004} \\
& & Dirichlet & \result{.014}{.005} & \result{.033}{.003} & \result{.024}{.002} & \result{.049}{.004} \\[3pt]
& \multirow{2}{*}{GMNet} & Softmax & \result{.131}{.020} & \result{.039}{.005} & \result{.041}{.005} & \result{.051}{.003} \\
& & Dirichlet & \result{.017}{.008} & \result{.033}{.002} & \result{.039}{.004} & \result{.052}{.004} \\
\bottomrule
\end{tabularx}
\end{table}

\subsection{Complete results for post-hoc prevalence scaling}
\label{app:scaling-results}

Table~\ref{tab:scaling-analysis-complete} reports the scaling
alternatives selected for the final symmetric configuration of each
dataset and sampling protocol. The comparison is based on validation
AE and is performed separately within each fold.

\begin{table}[t]
  \centering \footnotesize \setlength{\tabcolsep}{4pt}
  \renewcommand{\arraystretch}{1.08}

  \caption{Post-hoc prevalence scaling selected for the final
  symmetric configurations. Selection counts indicate the number of
  folds in which each alternative achieves the lowest validation
  AE. Boldface identifies the most frequently selected alternative for
  each dataset. The final column reports mean paired $\Delta$AE
  (standard deviation) across the five folds, computed as selected
  minus unscaled validation AE; negative values favour scaling.}
  \label{tab:scaling-analysis-complete}

\begin{tabularx}{\textwidth}{
 @{}ll
 >{\raggedright\arraybackslash}X
 >{\centering\arraybackslash}c
 >{\centering\arraybackslash}c
 >{\centering\arraybackslash}c
 >{\centering\arraybackslash}p{0.14\textwidth}
 @{}
}
\toprule \textbf{Protocol} & \textbf{Dataset} & \textbf{Selected
configuration} & \textbf{None} & \textbf{Scalar} & \textbf{Vector} &
$\boldsymbol{\Delta}\textbf{AE}$
\\
\midrule

\multirow{4}{*}{APP} & Cora-2 & GCN--GMNet--Dirichlet & \textbf{3/5} &
1/5 & 1/5 & \result{-.0001}{.0001}
\\

& Toloker & GESN--DeepSets--Dirichlet & 1/5 & \textbf{4/5} & 0/5 &
\result{-.0019}{.0014}
\\

& Cora & GCN--GMNet--Dirichlet & \textbf{2/5} & 1/5 & \textbf{2/5} &
\result{-.0024}{.0034}
\\

& Amazon & GESN--GMNet--Dirichlet & 2/5 & 0/5 & \textbf{3/5} &
\result{-.0037}{.0045}
\\

\midrule

\multirow{4}{*}{BFP} & Cora-2 & GCN--DeepSets--Dirichlet &
\textbf{2/5} & 1/5 & \textbf{2/5} & \result{-.0021}{.0030}
\\

& Toloker & GESN--GMNet--Dirichlet & \textbf{3/5} & 2/5 & 0/5 &
\result{-.0005}{.0008}
\\

& Cora & GCN--DeepSets--Softmax & \textbf{3/5} & 2/5 & 0/5 &
\result{-.0001}{.0002}
\\

& Amazon & GESN--DeepSets--Softmax & \textbf{5/5} & 0/5 & 0/5 &
\result{.0000}{.0000}
\\

\midrule

\multirow{4}{*}{RWP} & Cora-2 & GCN--GMNet--Dirichlet & \textbf{3/5} &
0/5 & 2/5 & \result{-.0003}{.0005}
\\

& Toloker & GESN--DeepSets--Softmax & \textbf{5/5} & 0/5 & 0/5 &
\result{.0000}{.0000}
\\

& Cora & GCN--DeepSets--Softmax & \textbf{4/5} & 1/5 & 0/5 &
\result{-.0000}{.0001}
\\

& Amazon & GESN--DeepSets--Softmax & \textbf{3/5} & 1/5 & 1/5 &
\result{-.0001}{.0003}
\\

\bottomrule
\end{tabularx}
\end{table}

\end{appendices}

\FloatBarrier


\bibliography{Domenico,Fabrizio}

\end{document}